\documentclass[letterpaper]{article} 
\usepackage[preprint]{aaai2027}
\usepackage[hyphens]{url}  
\usepackage{graphicx} 
\def\UrlFont{\rm}  
\usepackage{natbib}  
\usepackage{caption} 
\usepackage{algorithm}
\usepackage{algorithmic}

\usepackage{amsmath}
\usepackage{amssymb}
\usepackage{amsthm}
\usepackage{booktabs}
\usepackage{multirow}

\renewcommand{\topfraction}{0.95}
\renewcommand{\dbltopfraction}{0.95}
\renewcommand{\textfraction}{0.05}
\renewcommand{\floatpagefraction}{0.9}
\renewcommand{\dblfloatpagefraction}{0.9}

\newcommand{\E}{\mathbb{E}}
\newcommand{\R}{\mathbb{R}}
\newcommand{\KL}{\mathrm{KL}}
\newcommand{\TV}{\mathrm{TV}}
\newcommand{\norm}[1]{\left\lVert #1 \right\rVert}
\newcommand{\Swall}{S}

\theoremstyle{plain}
\newtheorem{theorem}{Theorem}
\newtheorem{lemma}{Lemma}
\newtheorem{proposition}{Proposition}
\newtheorem{corollary}{Corollary}
\theoremstyle{definition}
\newtheorem{definition}{Definition}
\newtheorem{assumption}{Assumption}
\newtheorem{remark}{Remark}

\title{Accelerating Time Series Foundation Models with Speculative Decoding}
\author{
Pranav Subbaraman$^{*1}$\quad
Fang Sun$^{*1}$\quad
Jinxi Yu$^{*1}$\quad
\\
Yue Yao$^{1}$\quad
Huacong Tang$^{1}$\quad
Xiao Luo$^{2}$\quad
Yizhou Sun$^{1}$
}
\affiliations{
$^{*}$Equal contribution, alphabetical order.\\
$^{1}$University of California, Los Angeles\quad
$^{2}$University of Wisconsin--Madison
}

\begin{document}

\maketitle

\begin{abstract}
Time series forecasting drives operational decisions under tight latency
budgets, and autoregressive time series foundation models (TSFMs)
increasingly deliver the most accurate forecasts. That accuracy is paid
for at inference, since a horizon of $H$ steps takes $\lceil H/P\rceil$
sequential forward passes of a large model, so latency grows with
exactly the long horizons these models are prized for. Yet a far
cheaper model predicts most next patches nearly as well as the large
one, and causal models can verify a block of future patches in one
parallel pass even though they generate them one at a time. These are
precisely the conditions under which speculative decoding thrives in
LLMs, but its ingredients are all defined over discrete vocabularies.
We therefore develop speculative decoding for continuous patch
autoregression. A cheap draft proposes $K$ future patches, and the
target verifies all of them in a single causal pass, accepting each by a
log-domain Gaussian likelihood-ratio test and correcting the first
rejection with its own prediction. We prove that the accelerated output
stays within a squared-error radius of target-only decoding set by an
acceptance temperature, and that throughput follows a capped-geometric
law that makes speedups predictable before deployment. The method
delivers up to $3.0\times$ inference speedup at accuracy between target
and draft across five TSFM families, and we characterize which
architectures admit single-pass verification and when speculation does
not pay.
\end{abstract}


\section{Introduction}

Time series forecasting sits on the critical path of operational
decisions. Electricity load, traffic control, cloud capacity, real-time
recommendation, and CDN provisioning all consume forecasts under strict
latency and throughput budgets
\citep{covington2016deep,nygren2010akamai}. Long
served by compact per-dataset models
\citep{salinas2020deepar,nie2023patchtst}, these workloads are shifting
to time series foundation models (TSFMs)
\citep{liu2025timerxl,das2024timesfm,ansari2024chronos,shi2024timemoe,%
liu2025sundial,auer2025tirex}, which are pretrained on billions of time
points, forecast unseen series zero-shot, and often beat supervised
baselines. Their accuracy, however, is bought with a decoding loop that
works against the workloads it serves. TSFMs forecast autoregressively,
each forward pass emitting one patch of $P$ future steps, so a horizon
of $H$ steps costs $\lceil H/P\rceil$ strictly sequential passes of the
full model. Latency thus grows with exactly the long horizons these
models are prized for, while deploying a smaller model surrenders the
accuracy that justified a foundation model.

\begin{figure}[t]
  \centering
  \includegraphics[width=0.94\columnwidth]{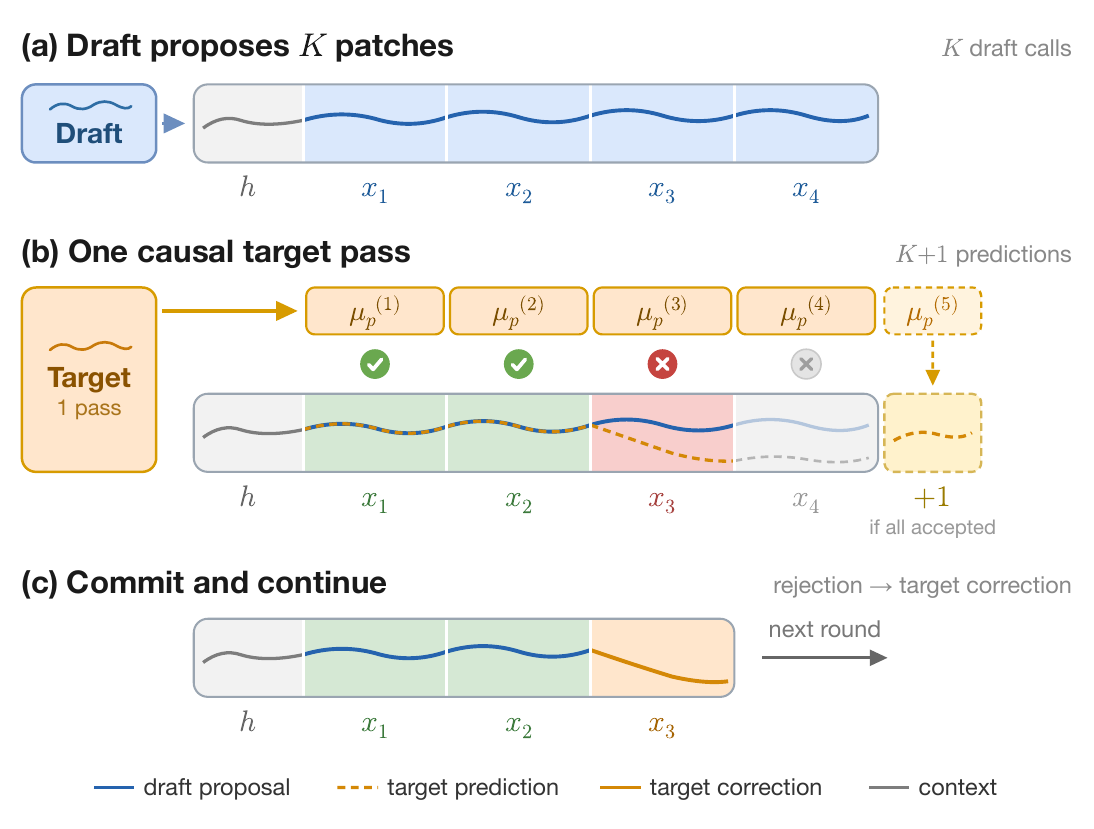}
  \caption{Speculative decoding for time series patches ($K{=}4$).
  (a)~The draft proposes $K$ patches beyond the history $h$.
  (b)~One causal target pass over $[h,x_1,\dots,x_K]$ predicts the next
  patch at all $K{+}1$ prefix boundaries; verifying left to right
  accepts $x_1,x_2$, rejects $x_3$, and discards later proposals
  untested (gray). (c)~The rejection is replaced by the target's own
  prediction $\mu_p^{(3)}$, so every round commits at least one patch;
  the $(K{+}1)$-st prediction is a free bonus patch (dashed) committed
  only on full acceptance.}
  \label{fig:overview}
\end{figure}

Two structural properties of forecasting show this bottleneck is not
inherent. First, forecasting difficulty is unevenly distributed along a
horizon. Recurring structure such as diurnal cycles and persistent
trends makes most next patches easy, so a far cheaper model predicts
nearly the same next patch as the large one on most steps, and the
large model earns its cost only on the hard transitions. Second, causal
models can \emph{verify} in parallel what they must \emph{generate}
sequentially, since one forward pass over the context plus a block of
tentative future patches scores every prefix at once. These are
precisely the conditions under which speculative decoding (SD) thrives
in large language models \citep{leviathan2023fast,chen2023speculative}.
There, a cheap draft proposes several steps, the target checks them all
in one pass, and sequential cost is paid only where the two disagree,
all without touching the deployed target's weights.

Transferring the idea is not mechanical, however, because every
ingredient of SD is defined over a discrete vocabulary, and three
questions must be answered. (1) \emph{What should the acceptance rule be
when outputs are continuous?} There are no token probabilities to
compare, and TSFM heads are heterogeneous (point, quantile, flow
matching), often exposing no density at all. We construct a log-domain
Gaussian likelihood-ratio test over patch densities, whose
mean-propagating form is the continuous analogue of greedy speculative
decoding. (2) \emph{What replaces the exact correction step?} Discrete
SD is lossless because rejections are resampled from a residual
distribution, but in $\R^d$ that draw costs a rejection loop whose
expected length diverges exactly in the high-acceptance regime where
speculation pays. We commit the target's own prediction instead.
(3) \emph{When does speculation actually save time?} Drafting adds
compute and verification widens the target pass, so the saved
sequential passes must outweigh the added parallel work. We derive a
throughput law whose inputs are measurable before deployment.

Our contributions are threefold. \emph{First}, we introduce a
speculative decoding framework operating natively on continuous patch
densities, removing the horizon-long sequential bottleneck of TSFM
decoding without retraining or modifying the target. \emph{Second}, we
develop the accompanying theory, comprising a capped-geometric
throughput law with a computable near-optimal block size, a provable
bound on the deviation from target-only decoding, and a closed-form
acceptance that makes speedups predictable before deployment.
\emph{Third}, we instantiate the framework on five TSFM families
(Timer-XL, TimesFM~2.5, Sundial, Time-MoE, TiRex) with drafts needing
no draft-specific pretraining. The framework delivers up to
$3.0\times$ speedups at accuracy between target and draft (acceptance
$49\%$--$97\%$), and we characterize which architectures structurally
admit single-pass verification and when speculation does not pay.

\section{Preliminaries}

\paragraph{Patch-Autoregressive Forecasting.}
Individual time points carry little information on their own, and
attention over point-level tokens is prohibitively long-ranged, so
modern forecasters tokenize a series into \emph{patches}, contiguous
windows of values that play the role of words \citep{nie2023patchtst}.
TSFMs adopt this interface for generation as well
\citep{liu2025timerxl,das2024timesfm}. We write $x_t\in\R^{d}$ for the
patch at step $t$, where $d=P\times C$ is the output patch length $P$
times the number $C$ of jointly modeled channels, and
$h_t=(h_0,x_{1:t-1})$ for the history
consisting of the context $h_0$ and all previously committed patches.
A decoder-only backbone defines the autoregressive factorization:
\begin{equation}
p(x_{1:T}\mid h_0)=\prod_{t=1}^{T}p(x_t\mid h_t),
\qquad T=\lceil H/P\rceil.
\label{eq:ar}
\end{equation}
Forecasting $H$ steps therefore commits $T$
patches in sequence, and each factor of Eq.~\eqref{eq:ar} costs one full
forward pass of the model.

\paragraph{Speculative Decoding.}
Speculative decoding descends from blockwise parallel decoding
\citep{stern2018blockwise}, which exploited the observation that a causal
Transformer can \emph{score} a block of tentative future tokens in one
forward pass even though it must \emph{generate} them one at a time.
\citet{leviathan2023fast} and \citet{chen2023speculative} turned this
into a lossless sampling accelerator for LLMs. A cheap draft model $q$
proposes $K$ tokens $x_1,\dots,x_K$ autoregressively, the target $p$
scores all $K{+}1$ prefixes in a single pass, and proposal $x_i$ is
accepted with probability:
\begin{equation}
\alpha_i=\min\Bigl\{1,\
\frac{p(x_i\mid h,x_{1:i-1})}{q(x_i\mid h,x_{1:i-1})}\Bigr\}.
\label{eq:lrrule}
\end{equation}
The first rejected position is replaced by a sample from the residual
distribution
\begin{equation}
\rho(x)\ \propto\ \bigl(p(x)-q(x)\bigr)_+,
\qquad (a)_+=\max\{a,0\},
\label{eq:residual}
\end{equation}
which makes the output provably distributed as if the target had
decoded alone, and when all $K$ proposals survive, the target's
$(K{+}1)$-st prediction is a free bonus token. Each round thus emits
between $1$ and $K{+}1$ tokens for one target call. The technique is
now standard in LLM serving, and a large literature refines the draft
side \citep{cai2024medusa,li2024eagle2,fu2024lookahead}.

\paragraph{From Tokens to Patches.}
Every ingredient above is defined over a discrete vocabulary, and each
breaks in the continuous setting. The acceptance rule in Eq.~\eqref{eq:lrrule}
divides two token probabilities, but a patch lives in $\R^d$ and
released TSFM heads often define no density at all. The correction step
requires sampling from the residual in Eq.~\eqref{eq:residual}, which in
$\R^d$ costs a rejection loop whose expected length diverges as target
and draft agree (Proposition~\ref{prop:rescost}), so exact correction
is most expensive precisely when speculation pays. And discrete
losslessness is the wrong goal anyway, since forecasts are consumed as
point predictions under squared error, where injected sampling noise is
pure harm. The next section rebuilds each ingredient for continuous
patches.

\section{Gaussian Speculative Decoding for Patches}

We rebuild the three ingredients in turn. A Gaussian surrogate
interface supplies densities over arbitrary continuous heads, a
tempered log-domain likelihood-ratio test replaces token-probability
acceptance, and a bounded-deviation fallback replaces residual
resampling. We then establish single-pass verification and the
guarantees each choice carries.

\subsection{Gaussian Surrogate Heads}

To obtain a common probabilistic interface across heterogeneous heads, we
equip target and draft with isotropic Gaussian \emph{surrogates} for
their next-patch conditionals,
\begin{equation}
p(x\mid h)=\mathcal{N}\!\bigl(\mu_p(h),\sigma_p^2 I_d\bigr),\quad
q(x\mid h)=\mathcal{N}\!\bigl(\mu_q(h),\sigma_q^2 I_d\bigr),
\label{eq:heads}
\end{equation}
where $\mu_p,\mu_q$ are the point predictions of target and draft,
namely the raw output for point-forecast heads (Timer-XL, Time-MoE),
the median for quantile heads (TimesFM, TiRex), and the sample mean for
the Sundial flow-matching head. The surrogate is instrumental rather
than a claim about a model's own predictive law, and the fidelity
guarantees below are statements about the surrogate chain. In all
families, $\sigma_p=\sigma_q=\sigma$ is a supplied modeling scale, the
\emph{acceptance temperature}, not a model output. Our scheme injects
\emph{no} proposal noise, and a learned draft variance enters only
Remark~\ref{rem:normterm}.

\subsection{Acceptance Rule}

For a proposal $x_i$ at position $i$, with target boundary prediction
$\mu_p^{(i)}$ and draft prediction $\mu_q^{(i)}$, define the per-dimension
energy $\norm{u}_d^2=\frac{1}{d}\norm{u}_2^2$. In direct analogy with
Eq.~\eqref{eq:lrrule}, the acceptance statistic is the $1/d$-scaled log
likelihood ratio $\ell_i=\tfrac{1}{d}\log[p(x_i)/q(x_i)]$ of the two
Gaussians in Eq.~\eqref{eq:heads},
\begin{equation}
\ell_i= -\frac{\norm{x_i-\mu_p^{(i)}}_d^2}{2\sigma_p^2}
  +\frac{\norm{x_i-\mu_q^{(i)}}_d^2}{2\sigma_q^2}
  +\frac{1}{2}\log\frac{\sigma_q^2}{\sigma_p^2},
\label{eq:logratio}
\end{equation}
and the proposal is accepted with probability
$\alpha_i=\min\{1,e^{\ell_i}\}$, computed in the log domain
with an overflow clamp. Under stochastic proposals and the unscaled
ratio, the mean acceptance equals the overlap $\int\min\{p,q\}$ of the
two densities. Two departures from the LLM rule are deliberate. The
$1/d$ scaling tempers the
ratio to $(p/q)^{1/d}$. Without it, the raw log ratio grows linearly in
$d$ and acceptance degenerates into a nearly $0/1$ gate, whereas
tempering keeps $\sigma$ an $O(1)$ acceptance temperature across
families (Remark~\ref{rem:dscaling}); Table~\ref{tab:ablation} measures
this. The last term of
Eq.~\eqref{eq:logratio}, the Gaussian normalization (log-determinant) term
absent from any discrete ratio, vanishes under the shared scale
$\sigma_p=\sigma_q$ that we deploy, and is load-bearing only when the
draft carries a scale of its own
(Remark~\ref{rem:normterm}); we therefore measure it in that separate
configuration rather than in Table~\ref{tab:ablation}. Acceptance is evaluated per example and per
position, left to right within a round, and the round ends at the first
rejection.

\subsection{Mean Propagation and Fallback}

\paragraph{Mean-Propagating Acceptance (Our Scheme).}
The draft proposes its mean, $x_i=\mu_q^{(i)}$, and the fallback commits the
target mean, $t=\mu_p^{(i)}$ ($r=0$). Substituting $x_i=\mu_q^{(i)}$ into
Eq.~\eqref{eq:logratio} with $\sigma_p=\sigma_q=\sigma$ collapses acceptance to a
distance gate,
\begin{equation}
\alpha_i=\exp\!\Bigl(-\frac{\norm{\mu_q^{(i)}-\mu_p^{(i)}}_d^2}{2\sigma^2}\Bigr),
\label{eq:detgate}
\end{equation}
so the draft patch is kept, with high probability, exactly when the target
would have predicted nearly the same patch. This is the continuous
analogue of \emph{greedy} (argmax) speculative decoding. Where the LLM
variant keeps a drafted token only if it matches the target's argmax,
Eq.~\eqref{eq:detgate} relaxes the exact match, which has measure zero in
$\R^d$, to a soft distance gate, and $\sigma\to0$ recovers exact-match
acceptance. Propagating conditional means rather than samples injects no
noise and suits squared-error evaluation, and provably keeps each
committed patch within a squared-error radius of the target's own forecast
(Proposition~\ref{prop:detfidelity}). It is the configuration used
throughout our experiments. Since the energy is normalized by
$d=P\times C$, $\sigma$ values are not comparable across families.

\paragraph{Fallback Correction and Sampled Variants.}
On rejection at position $i$ the round commits the target mean
$t=\mu_p^{(i)}$. More generally it may commit a draw
$t=\mu_p^{(i)}+r\,\varepsilon$ from $p_r=\mathcal{N}(\mu_p^{(i)},r^2I_d)$ with
fallback noise $r\le\sigma_p$, where $r=\sigma_p$ draws from the target
law itself. The same acceptance rule equally admits sampled proposals
$x_i=\mu_q^{(i)}+\sigma_q\,\varepsilon$. Both noise knobs are exercised
only in the ablation study, since sampling adds error under
squared-error evaluation. We do not perform residual resampling from
Eq.~\eqref{eq:residual}, and the guarantees below explain why its exactness
is not worth its cost here.

\paragraph{Bonus Patch on Full Acceptance.}
When all $K$ proposals of an example are accepted, the $(K{+}1)$-st boundary
prediction from the same verification pass is committed (as the target mean,
or with fallback noise $r$ when noise is enabled), so a fully accepted round
emits $K{+}1$ patches with no extra target call.

\subsection{Single-Pass Verification}

One verification round runs the target once on the concatenation
$[h,\,x_1,\dots,x_K]$ under the model's causal mask and reads the next-patch
prediction at the $K{+}1$ prefix boundaries (Figure~\ref{fig:overview}).
Two details matter in practice. First, \emph{normalization is frozen on
the context}, meaning instance statistics come from $h$ only, since
proposals would otherwise leak into the normalization of their own
verification. Second, \emph{boundary indexing is family-specific}.
TiRex does not expose per-position
next-patch outputs, so its verification stacks the $K$ prefixes into
one batched call, which remains a single call per round but with
compute growing in $K$.

\paragraph{Architectural Requirements.}
Single-pass verification requires (i) causal autoregression, so that one
forward over context plus proposals scores all prefixes, and (ii) a
continuous per-patch output with a usable point or density summary.
None needs
draft-specific pretraining, as only Timer-XL trains a small draft and
the rest reuse existing checkpoints. Chronos-Bolt \citep{ansari2024chronos} and
Moirai \citep{woo2024moirai} fail (i)
structurally, as do the roughly $35$ supervised direct multi-horizon
forecasters in the Time Series Library collection \citep{wu2023timesnet}.
Chronos-T5 is causal but emits discrete tokens. Applying the Gaussian
rule to its dequantized samples degraded MSE monotonically with
acceptance, up to $37\%$ above target in an ETTh1 check, so token-level
(LLM) speculative decoding is the appropriate tool there.

\subsection{Theoretical Guarantees}

A practitioner adopting the scheme needs to know how much faster it is
and how much of the target's accuracy is at risk. The throughput law
holds for \emph{any} acceptance rule, since speed depends only on how
often proposals are accepted. The fidelity guarantee bounds how far the
accelerated output can drift from target-only decoding, licensing trust
wherever the target's forecasts were trusted. Supporting results and
all proofs are in the technical appendix, submitted separately;
lettered numbers such as Proposition~\ref{prop:kernel} or
Table~\ref{tab:ksweep} refer to it.

\paragraph{Throughput and Block-Size Choice.}
Let $A_i$ be the event that position $i$ of a round is accepted and suppose
$\Pr(A_i\mid A_1,\dots,A_{i-1})=\bar\alpha$ within a round. The number of
patches $L\in\{1,\dots,K{+}1\}$ emitted per round follows a capped geometric
law,
\begin{equation}
\E[L]=\frac{1-\bar\alpha^{K+1}}{1-\bar\alpha},
\label{eq:EL}
\end{equation}
a capped geometric tail sum (Proposition~\ref{prop:blocklength}, with
Proposition~\ref{prop:dependence} handling dependent acceptance
events). With $c$ the draft-to-target per-call
GPU cost ratio, $v$ the verification-to-forward cost ratio, and the bonus
patch free, the GPU-compute speedup of one round over target-only decoding
is
\begin{equation}
\Swall(K)=\frac{\E[L]}{cK+v}
=\frac{1-\bar\alpha^{K+1}}{(1-\bar\alpha)(cK+v)}.
\label{eq:speedup}
\end{equation}
These expressions depend only on acceptance events, not on how proposals
are formed, so they hold whether or not noise is injected.
Proposition~\ref{prop:kstar} gives the marginal criterion
$\Swall(K{+}1)\ge\Swall(K)$ iff $\bar\alpha^{K+1}(cK+v)\ge c\,\E[L_K]$,
so a near-optimal block size $K^\star$ is computable from the measured
pair $(\hat{\bar\alpha},c)$. Both inputs are estimable before
deployment. For equal covariances, $\bar\alpha=2\Phi(-\Delta/2)$ in the
Mahalanobis gap $\Delta$, and its plug-in estimate from held-out
histories concentrates at Hoeffding rate
(Proposition~\ref{prop:closedform}, Theorem~\ref{thm:hoeffding}), so
Eqs.~\eqref{eq:EL}--\eqref{eq:speedup} are predictable in advance.

\paragraph{Fidelity to Target-Only Decoding.}
Speed is worthless at unbounded accuracy cost, so we bound how far the
committed output can stray from what the target would have decoded
alone. Our accelerator is deliberately approximate. Residual resampling
would make it exact (Theorem~\ref{thm:block}), but only relative to the
surrogate of Eq.~\eqref{eq:heads}, and its rejection loop takes $1/(1-\beta)$
expected target draws (Proposition~\ref{prop:rescost}), a factor of
$3$--$20$ at the overlaps $\beta\in[0.7,0.95]$ we operate at. Exactness
therefore costs the most precisely when speculation helps the most.
Falling back to the target's own prediction instead keeps the gap
provably small. For stochastic proposals the one-step output law is
$g=\alpha q+(1-\bar\alpha)p$, whose total-variation distance to $p$ is
at most $1-\bar\alpha$ under the likelihood-ratio rule, and per-step
bounds compose additively over the horizon
(Proposition~\ref{prop:kernel}, Lemma~\ref{lem:fallback}). For our
mean-propagating scheme the right metric is squared error, and the
guarantee is the following.

\begin{proposition}[Deterministic fidelity to the target forecast]
\label{prop:detfidelity}
Fix a position with prefix $h$ and per-dimension draft--target gap
$\Delta^2=\norm{\mu_q-\mu_p}_d^2$. Under the deterministic gate of
Eq.~\eqref{eq:detgate} with acceptance temperature $\sigma$ and target-mean
fallback ($r=0$), the committed patch $Y$ (equal to $\mu_q$ on acceptance,
$\mu_p$ on rejection) satisfies
\begin{equation}
\E\bigl[\norm{Y-\mu_p}_d^2\mid h\bigr]
=\Delta^2 e^{-\Delta^2/2\sigma^2}\le\frac{2\sigma^2}{e},
\label{eq:detfidelity}
\end{equation}
a bound independent of the gap, with tail
$\Pr\bigl(\norm{Y-\mu_p}_d^2>\tau\mid h\bigr)\le e^{-\tau/2\sigma^2}$. Thus
$\sigma$ is a per-patch fidelity radius to target-only decoding under
squared error.
\end{proposition}

The scheme moreover commits only $\mu_q$ or $\mu_p$, the Bayes actions
under squared error, so injecting noise can only inflate error
(Remark~\ref{rem:detpoint}), as the ablation of Figure~\ref{fig:noise}
confirms. The one silent failure mode we found is also provably
diagnosed. With a learned-variance draft ($\sigma_q\ll\sigma_p$),
omitting the determinant term of Eq.~\eqref{eq:logratio} inflates the
statistic by $\log(\sigma_p/\sigma_q)>0$, a factor $d$ larger in the
untempered log ratio, and drives $\alpha\to1$
regardless of draft--target agreement, silently turning speculative
decoding into draft-only decoding (Remark~\ref{rem:normterm}).

\paragraph{Predictable Failure Conditions.}
The same quantities delimit, before any deployment, where speculation
cannot pay. A precondition is that the target be worth waiting for.
When the draft already matches the target's accuracy there is nothing
to accelerate and draft-only decoding dominates, a condition we call
the \emph{accuracy gate} and test directly. Past the gate,
Eq.~\eqref{eq:speedup} exceeds one only if $\E[L]>cK+v$, which fails in
three measurable ways. Acceptance $\bar\alpha$ can be too low at every
accuracy-preserving temperature, so that $\E[L]$ stays near one. The
draft cost $c$ can be too large a fraction of the target's for the
acceptance achieved. And the verification cost $v$ can grow with $K$
instead of staying near one (Remark~\ref{rem:verifycost}). All three
quantities are estimable from held-out batches
(Theorem~\ref{thm:hoeffding}, Remark~\ref{rem:estimating}), so these
failures are predictable rather than discovered in production. The
failure analysis in the experiments matches each observed failure to
one of these conditions.

\section{Experiments}

Our experiments ask whether Gaussian speculative decoding delivers real
speedups at acceptable accuracy across model families, how the operating
point responds to block size, batch size, draft configuration, and
horizon, which design choices carry its performance, how accelerated
forecasts behave on concrete series, and where speculation fails to pay.
Every comparison is a matched three-way run of target-only, draft-only,
and speculative decoding, and negative results are reported alongside
wins.

\subsection{Setup}

\paragraph{Models and Drafts.}
Timer-XL (67.4M parameters, context $1536$, patch $96$) is drafted by a
$0.125\times$-width variant (158K parameters, $0.23\%$ of the target),
distilled from the target on ETTh1 and trained supervised on the other
datasets. TimesFM 2.5 (231.3M loaded) is drafted by the
same checkpoint at context $256$ with the quantile head disabled on the
draft and flip invariance off in all three modes. Sundial-base (128.3M)
runs $20$ flow samples as target and is drafted by itself at $1$ sample
and context $512$. Time-MoE-200M (453.2M loaded, context $3072$) is
drafted by Time-MoE-50M (113.4M, context $768$). TiRex (35M) emits $m$
$32$-step patches per forward, with the target decoding at $m{=}4$, the
draft at $m{=}16$, and each draft rollout verified as $K{=}4$ blocks of
four patches. Timer-XL uses supervised per-dataset checkpoints, and the
other four families are zero-shot.

\begin{table*}[tp]
  \centering
  \small
  \setlength{\tabcolsep}{5pt}
  \begin{tabular}{@{}llrrrrrrrr@{}}
  \toprule
  Model & Dataset & $B$ & $K$ & $\sigma$ & Accept & Target & Draft & Spec. & Speedup \\
  \midrule
  \multirow{11}{*}{Timer-XL} & ECL & 16 & 1 & 0.25 & 68.7\% & 0.170 & 0.229 & 0.193 & 1.29$\times$ \\
   & ECL & 1 & 3 & 0.25 & 76.8\% & 0.120 & 0.163 & 0.151 & 1.53$\times$ \\
   & ECL ($H{=}720$) & 16 & 3 & 0.25 & 78.3\% & 0.197 & 0.278 & 0.245 & 1.67$\times$ \\
   & ECL ($H{=}720$) & 1 & 3 & 0.25 & 81.5\% & 0.200 & 0.258 & 0.240 & \textbf{1.82$\times$} \\
   & Weather & 1 & 3 & 0.25 & 65.0\% & 0.293 & 0.366 & 0.332 & 1.16$\times$ \\
   & Weather ($H{=}720$) & 1 & 3 & 0.25 & 69.2\% & 0.320 & 0.399 & 0.362 & 1.42$\times$ \\
   & PEMS03 & 1 & 3 & 0.25 & 53.9\% & 0.062 & 0.200 & 0.113 & 1.17$\times$ \\
   & PEMS04 & 1 & 3 & 0.35 & 57.8\% & 0.091 & 0.341 & 0.214 & 1.16$\times$ \\
   & PEMS08 & 1 & 3 & 0.35 & 55.5\% & 0.164 & 0.360 & 0.261 & 1.22$\times$ \\
   & Traffic & 1 & 3 & 0.25 & 49.4\% & 0.300 & 0.444 & 0.350 & 1.04$\times$ \\
   & ETTh1 & 64 & 1 & 0.10 & 0.7\% & 0.475 & 0.547 & 0.476 & \textit{0.82$\times$} \\
  \midrule
  \multirow{6}{*}{TimesFM} & ETTh1 & 8 & 1 & 0.30 & 93.5\% & 0.489 & 0.524 & 0.496 & 1.39$\times$ \\
   & ETTh2 & 8 & 1 & 0.30 & 95.6\% & 0.428 & 0.521 & 0.456 & 1.45$\times$ \\
   & ETTm2 & 8 & 1 & 0.30 & 93.8\% & 0.444 & 0.530 & 0.472 & 1.43$\times$ \\
   & Weather & 8 & 1 & 0.30 & 95.2\% & 0.474 & 0.734 & 0.487 & 1.45$\times$ \\
   & ECL & 8 & 1 & 0.30 & 93.7\% & 0.300 & 0.342 & 0.314 & \textbf{1.50$\times$} \\
   & Traffic & 4 & 1 & 0.30 & 87.4\% & 0.451 & 0.532 & 0.479 & 1.43$\times$ \\
  \midrule
  \multirow{11}{*}{Sundial} & ETTh1 & 8 & 1 & 0.15 & 63.8\% & 0.378 & 0.452 & 0.388 & \textbf{1.38$\times$} \\
   & ETTh2 & 128 & 1 & 0.20 & 90.2\% & 0.312 & 0.340 & \textbf{0.311} & 1.19$\times$ \\
   & ETTm1 & 128 & 1 & 0.20 & 85.8\% & 0.270 & 0.319 & 0.280 & 1.17$\times$ \\
   & ETTm2 & 128 & 1 & 0.20 & 94.8\% & 0.064 & 0.076 & 0.068 & 1.23$\times$ \\
   & ECL & 4 & 1 & 0.30 & 92.5\% & 0.069 & 0.100 & 0.078 & 1.06$\times$ \\
   & Solar & 8 & 1 & 0.30 & 87.4\% & 0.200 & 0.466 & 0.264 & 1.02$\times$ \\
   & PEMS03 & 4 & 1 & 0.30 & 91.2\% & 0.097 & 0.170 & 0.114 & 1.08$\times$ \\
   & PEMS04 & 4 & 1 & 0.25 & 96.2\% & 0.136 & 0.166 & \textbf{0.128} & 1.09$\times$ \\
   & PEMS07 & 4 & 1 & 0.30 & 95.8\% & 0.180 & 0.185 & \textbf{0.171} & 1.14$\times$ \\
   & PEMS08 & 4 & 1 & 0.30 & 96.9\% & 0.264 & 0.433 & 0.330 & 1.15$\times$ \\
   & Traffic & 4 & 1 & 0.30 & 69.0\% & 0.381 & 0.481 & 0.403 & \textit{0.99$\times$} \\
  \midrule
  \multirow{7}{*}{Time-MoE} & Weather & 128 & 3 & 0.25 & 96.9\% & 0.405 & 0.438 & 0.410 & 2.59$\times$ \\
   & ETTm1 & 128 & 3 & 0.30 & 93.5\% & 0.430 & 0.440 & \textbf{0.415} & \textbf{3.05$\times$} \\
   & ECL & 16 & 3 & 0.15 & 91.1\% & 0.357 & 0.540 & 0.447 & 2.78$\times$ \\
   & PEMS03 & 16 & 3 & 0.15 & 76.4\% & 0.218 & 0.264 & 0.220 & 2.20$\times$ \\
   & PEMS07 & 8 & 3 & 0.15 & 72.3\% & 0.239 & 0.318 & 0.268 & 1.74$\times$ \\
   & Traffic & 8 & 3 & 0.15 & 72.5\% & 0.424 & 0.574 & 0.476 & 2.02$\times$ \\
   & Solar & 32 & 3 & 0.15 & 63.1\% & 0.526 & 0.764 & \textbf{0.497} & 1.75$\times$ \\
  \midrule
  \multirow{3}{*}{TiRex} & ETTh1 & 4 & 4 & 0.15 & 76.8\% & 0.623 & 0.743 & 0.624 & \textbf{1.56$\times$} \\
   & ETTh2 & 4 & 4 & 0.05 & 61.9\% & 0.547 & 0.647 & \textbf{0.538} & 1.18$\times$ \\
   & PEMS07 & 4 & 4 & 0.15 & 27.3\% & 0.191 & 0.360 & \textbf{0.191} & \textit{0.35$\times$} \\
  \bottomrule
  \end{tabular}
  \caption{One valid operating point per reported family--dataset
  configuration under the
  mean-propagating scheme (MSE and GPU compute time against the matched
  target-only baseline), selected from each pair's $\sigma$ sweep as the
  largest-speedup point whose speculative MSE lies below the
  target--draft midpoint, else a point faster than target-only and more
  accurate than draft-only where one exists. \textbf{Bold} Spec.\ marks
  speculative MSE at or below the target's, \textbf{bold} Speedup the
  best per family, and \textit{italic} Speedup slower than target-only.}
  \label{tab:main}
\end{table*}

\paragraph{Datasets.}
We evaluate on the standard long-horizon forecasting benchmarks, chosen
to span energy, weather, and transportation and to cover
channel counts from $7$ to $883$, so that acceptance is tested across
regimes rather than on a single domain:
the four ETT electricity-transformer datasets, Weather, ECL, Traffic,
Solar, and the PEMS traffic-flow collection (PEMS03/04/07/08). We
use the chronological train/validation/test splits standard in
long-term forecasting
\citep{lai2018lstnet,wu2021autoformer,liu2022scinet,liu2025timerxl},
and all metrics are computed on the test split.

\paragraph{Evaluation Protocol.}
Every comparison reports the three modes (target-only, draft-only, and
speculative) under identical dataset, split, batch size, precision,
context and prediction lengths, and hardware (a single H200).
Prediction lengths are $336$ steps for Timer-XL and Sundial ($720$
where stated) and $1536$ for TimesFM, Time-MoE, and TiRex. We report
MSE on the test splits, empirical
acceptance as the accepted fraction of proposals, and GPU compute time
per batch from CUDA events around model forwards (draft, verification,
and bonus sampling), after per-family warm-up batches and with fixed
seeds. Python-side logic and data loading
are excluded from all modes. Speedup is target-only over speculative
GPU time under matched settings, and for every family the baseline and
the verification pass run the same forward implementation (mask
handling, channel batching, and sampling), so $v$ in Eq.~\eqref{eq:speedup}
is not distorted. Headline runs propagate means with $\sigma$ swept per
dataset and no acceptance bias. As an end-to-end correctness check,
$\sigma\to0$ recovers target-only outputs exactly (acceptance $0$).
Per-family configurations, decoding pseudocode, dataset statistics, and
the complete sweep tables are in the technical appendix.

\subsection{Main Results}

\begin{figure*}[t]
\centering
\includegraphics[width=0.95\textwidth]{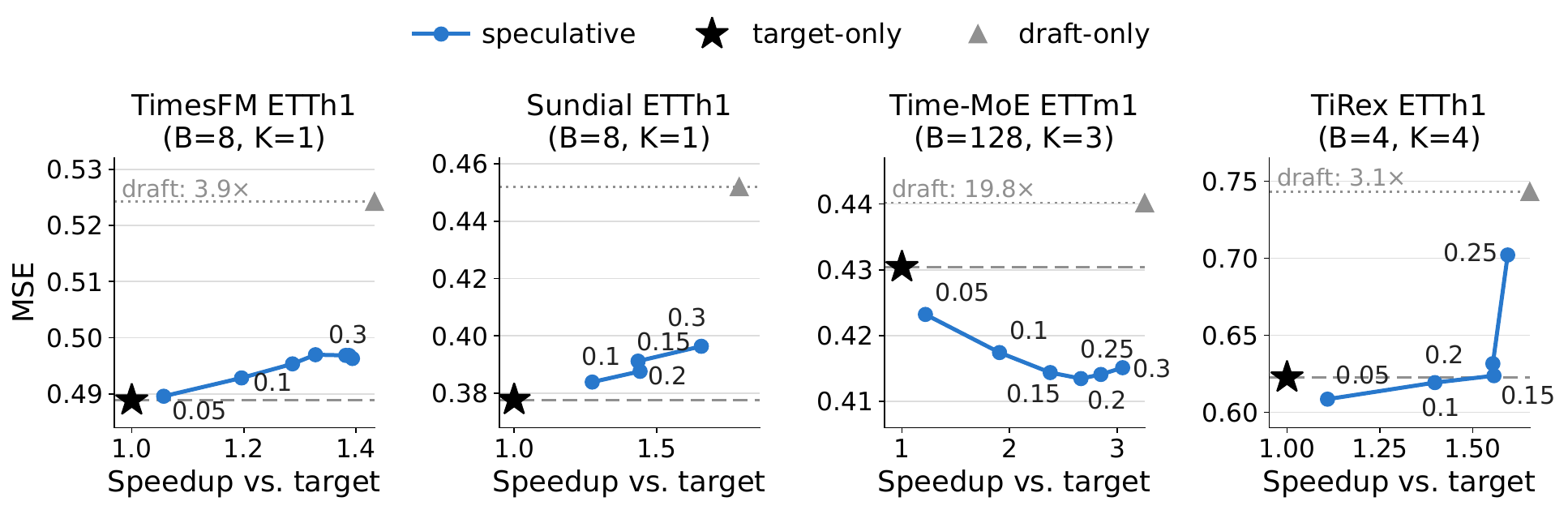}
\caption{Accuracy--speed frontier as $\sigma$ (point labels) is swept.
Dashed lines mark target-only MSE, dotted lines draft-only MSE, and the
grey triangle a draft speedup beyond the axis range. Speedups are
GPU-compute ratios against the same run's target-only baseline. Panels
show TimesFM and Sundial on ETTh1, Time-MoE on ETTm1, and TiRex on
ETTh1 (batch 4).}
\label{fig:tradeoff}
\end{figure*}

\begin{figure}[tb]
\centering
\includegraphics[width=1.00\columnwidth]{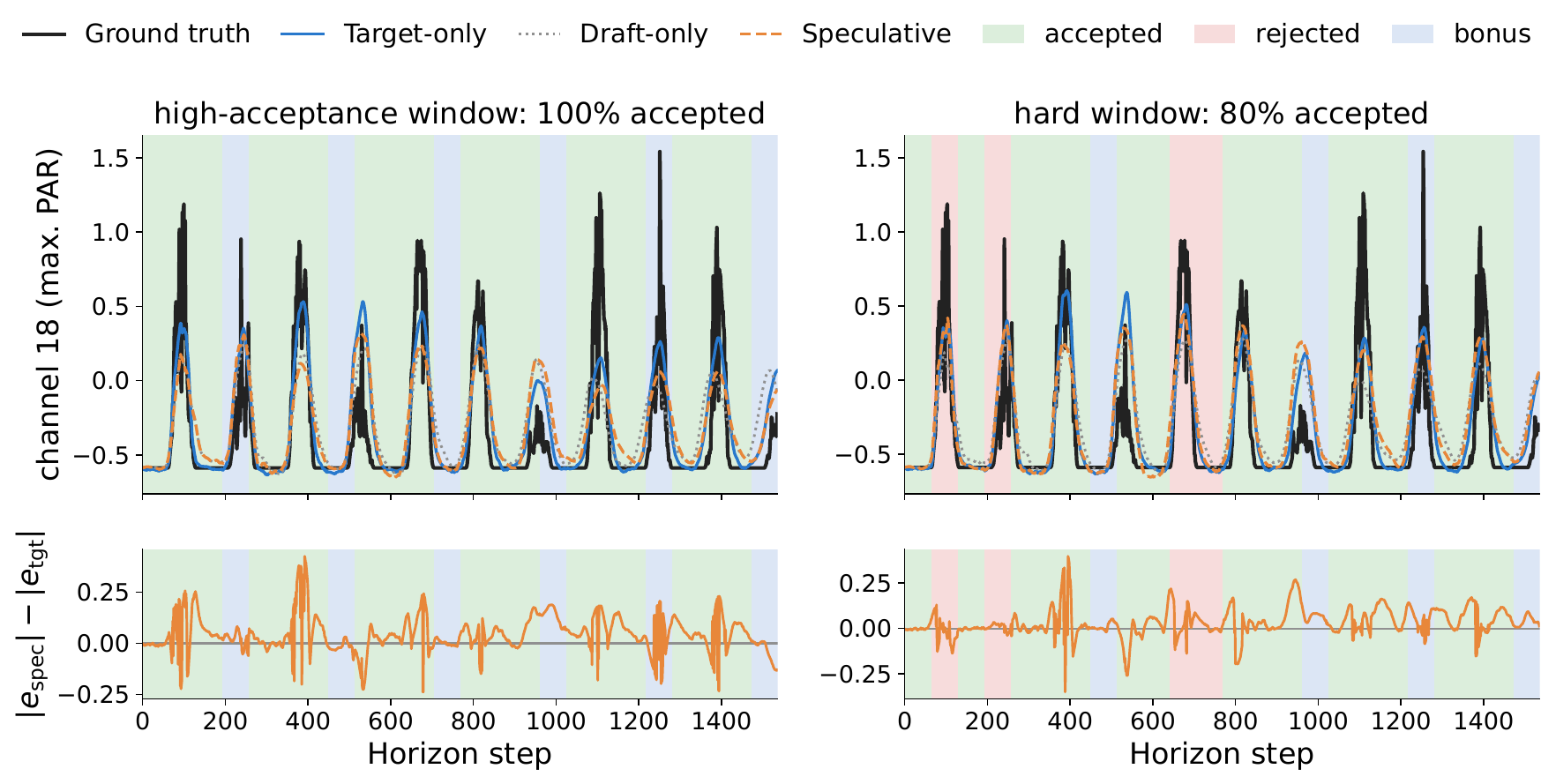}
\caption{Case study: Time-MoE on Weather ($B{=}8$, $K{=}3$,
$\sigma{=}0.15$), channel $18$. Top: target-only, draft-only, and
speculative forecasts against the ground truth, with each committed
patch shaded by provenance (accepted proposal, rejection fallback, or
the free bonus patch of a fully accepted round). Bottom: the pointwise
error difference to the target-only forecast. Left is the
highest-acceptance window of the set, right the lowest. The bounded
$64$-window subset is its own reference point and is not comparable with
Tables~\ref{tab:main} and~\ref{tab:ksweep}.}
\label{fig:case}
\end{figure}

Speculative decoding accelerates every family we instantiate it on. Of
the $38$ operating points in Table~\ref{tab:main}, one per reported
family--dataset configuration and selected from its $\sigma$ sweep under
the caption's criterion, $35$ beat target-only decoding, with family-best
speedups of $1.38\times$--$3.05\times$, and in seven rows the
accelerated forecasts are \emph{more accurate} than the target itself
(bold Spec.\ entries). Three observations hold across families. First,
$\sigma$ traces a monotone frontier from target-like accuracy at low
acceptance to draft-like accuracy at high acceptance, so a single knob
sets the accuracy--speed trade-off. Under mean propagation the entire
swept range stays below the draft line at a speedup above one for the
TimesFM and Time-MoE sweeps of Figure~\ref{fig:tradeoff}.
Second, the largest speedups come from the largest target-to-draft cost
ratios. Time-MoE (4:1 parameter and context ratios) reaches
$3.05\times$ at $93\%$ acceptance on ETTm1, whereas TimesFM's
short-context self-draft saturates near $1.5\times$ because the draft
still costs about a quarter of the target. Third, speculative decoding
can beat the target under squared error. A plausible mechanism, which
we report as a hypothesis, is reduced error accumulation. Target-only
decoding free-runs on its own rollout for tens of sequential patches,
while each rejection re-anchors the chain with the target's single-step
prediction. Consistently, the clearest case is Time-MoE at the longest
horizon ($H{=}1536$), where ETTm1's entire $\sigma$ sweep sits below
the target (Figure~\ref{fig:tradeoff}) and Solar is below target at the
operating point as well. Figure~\ref{fig:case} gives a representative
trajectory, in which acceptance climbs along the horizon and each
rejection pulls the chain back toward the target-only rollout.

\paragraph{Deployment Sensitivity.}
Full block-size, batch-size, draft-configuration, and horizon sweeps
(Tables~\ref{tab:ksweep} and~\ref{tab:bsweep}) confirm the predicted
acceptance--cost trade-off: the best $K$ and batch size are family- and
hardware-dependent, while longer horizons strengthen speculation.

\subsection{Ablation Studies}

Four choices distinguish the scheme from a naive transplant: what is
propagated (means rather than samples), what a rejection commits (the
target mean rather than a resample), what enters the acceptance
statistic (the $1/d$ tempering), and how the $K$ proposals are verified
(one pass rather than $K$ re-encodings). We ablate each in isolation at
the deployed configuration.

\paragraph{Mean Propagation versus Sampled Proposals.}
On Timer-XL ETTh1 (batch $64$, $K{=}1$, shared baseline) we sweep
injected proposal noise against noise-matched target-only and draft-only
baselines, and separately sweep fallback noise $r$ (four-arm protocol
and plots in Figure~\ref{fig:noise}). MSE
degrades monotonically with proposal noise while acceptance rises to
$94\%$, yet at every noise level the speculative chain stays at or below
the noise-matched target and far below the noise-matched draft
(Proposition~\ref{prop:kernel}). Fallback noise only hurts
(Lemma~\ref{lem:fallback}), with MSE rising from $0.4877$ at $r{=}0$ to
$1.407$ at $r{=}1.0$; Sundial (ETTh1, $n_s{=}10$) tracks its
noise-matched target within $0.013$ MSE.
Propagating means therefore dominates for point forecasts.

\paragraph{Component Ablation.}
Table~\ref{tab:ablation} separates two roles: the acceptance rule
decides how much accuracy the scheme keeps, the verifier how much time
it saves. Dropping the $1/d$ tempering collapses acceptance to $0.0\%$,
the $0/1$ gate of Remark~\ref{rem:dscaling}, leaving target-only
decoding at $32\%$ more cost since the draft is still paid for.
Sampling proposals and resampling on rejection leave acceptance within
$3.5$ points of the reference and cost only accuracy ($+15.3\%$ and
$+8.6\%$ MSE). Replacing
single-pass verification by the naive $K$-batched re-encoding changes
acceptance and accuracy only marginally ($70.2\%$ against $68.4\%$,
$0.281$ against $0.285$, from the naive verifier's per-window
renormalization) but turns $1.47\times$ into
$0.59\times$: the speed benefit comes entirely from $O(1)$
verification.

The normalization term vanishes under the shared scale
$\sigma_p=\sigma_q$, so it cannot be removed at this operating point.
Where it is live, dropping it raises acceptance from $61.4\%$ to
$90.1\%$ and degenerates the chain to draft-only decoding
(Remark~\ref{rem:normterm}): silently on Weather, where the draft is
nearly as accurate as the target ($0.282$ to $0.284$ MSE), but not on
ECL, where a far weaker draft sends MSE from $0.349$ to $0.897$ while
the broken variant reports the \emph{higher} speedup.

\subsection{Failure Analysis}

Every failure we observed realizes one of the conditions predicted in
the Theoretical Guarantees, and each is detectable cheaply before
deployment.

\paragraph{Accuracy Gate.}
Speculative decoding presumes the target is worth waiting for. On $12$
family--dataset pairs the draft matched or beat the target's MSE (the
list is in the appendix), so draft-only decoding
dominates and we skip speculative runs by protocol. For the supervised
Timer-XL pairs this reflects the checkpoint pair rather than the
algorithm.

\paragraph{Cost Structure.}
Even with the gate passed, speculative decoding loses when the
throughput condition $\E[L]>cK+v$ fails. Verification compute scaling
with $K$ produces the TiRex failure on PEMS07 (acceptance $27\%$,
$0.35\times$). An accuracy-preserving $\sigma$ region with near-zero
acceptance produces the Timer-XL failures on PEMS04, PEMS08, and
Traffic at batch ${\ge}4$. A draft that is not cheap combined with only
moderate acceptance produces the Sundial result on Traffic ($69\%$
acceptance at the fastest sweep point, $0.99\times$). In each case
Eq.~\eqref{eq:speedup} predicts the failure from the measured
$(\bar\alpha,c,v)$, so none need be discovered in production.

\section{Related Work}

\paragraph{Time Series Forecasting Models.}
Deep forecasters
\citep{salinas2020deepar,oreshkin2020nbeats,lai2018lstnet,liu2022scinet,
zhou2021informer,wu2021autoformer,liu2024itransformer} and the patch
tokenization popularized by PatchTST \citep{nie2023patchtst} led to
pretrained foundation models
\citep{liu2024timer,liu2025timerxl,das2024timesfm,ansari2024chronos,
goswami2024moment,woo2024moirai,liu2024moiraimoe,shi2024timemoe,
liu2025sundial,auer2025tirex}. This line inherits patch-by-patch
autoregression and its sequential inference cost. We treat these models
as fixed targets and change only their decoding loop.

\begin{table}[tb]
  \centering
  \small
  \setlength{\tabcolsep}{4pt}
  \begin{tabular}{@{}lrrr@{}}
  \toprule
  Configuration & Accept & MSE & Speedup \\
  \midrule
  Target only & --- & $0.277$ & $1.00\times$ \\
  \midrule
  Full rule (ours) & $68.4\%$ & $0.285$ & $1.47\times$ \\
  w/o $1/d$ tempering & $0.0\%$ & $0.277$ & $0.76\times$ \\
  w/o mean proposal ($x_i\!\sim\!q$) & $64.9\%$ & $0.329$ & $1.41\times$ \\
  w/o clean fallback ($r{=}\sigma$) & $67.1\%$ & $0.310$ & $1.44\times$ \\
  w/o single-pass verification & $70.2\%$ & $0.281$ & $0.59\times$ \\
  \bottomrule
  \end{tabular}
  \caption{Component ablation of the deployed configuration: Timer-XL
  Weather, $8$-layer target with the supervised $0.125\times$ draft,
  $K{=}3$, $B{=}64$, $\sigma{=}0.25$, horizon $336$, full test split
  ($10{,}204$ windows), all runs consecutive on one idle GPU. Each row
  replaces one component. The last row's MSE dips below the full rule
  because the naive verifier renormalizes each re-encoded input on its own
  window; its load-bearing column is the speedup.}
  \label{tab:ablation}
\end{table}

\paragraph{Speculative Decoding.}
Blockwise parallel decoding \citep{stern2018blockwise} first exploited
parallel verification of drafted continuations, accepting a token only
on an exact argmax match. \citet{leviathan2023fast} and
\citet{chen2023speculative} generalized acceptance to the token
probability ratio, making the accelerator provably lossless for sampled
decoding, and \citet{sun2023spectr} recast that rule as optimal
transport. Later variants improve the draft side
\citep{cai2024medusa,li2024eagle2,fu2024lookahead} or self-draft from
the target's own shallower computation
\citep{zhang2024draftverify,elhoushi2024layerskip}, as three of our
families do. All define
acceptance, correction, and losslessness over a finite vocabulary.
Continuous-output extensions remain limited to scalar or compact outputs
with exposed densities and purpose-built or closed-form decoders
\citep{wang2024continuous,gong2025tppsd,bilos2025speculative}, rather
than heterogeneous high-dimensional TSFM patches. Our
mean-propagating scheme is the continuous analogue of the greedy
variant, relaxing the argmax match, of measure zero in $\R^d$, to the
distance gate in Eq.~\eqref{eq:detgate} and replacing losslessness with the
bounded deviation of Proposition~\ref{prop:detfidelity}. Orthogonal
kernel- and serving-level optimizations
\citep{dao2022flashattention,ainslie2023gqa,kwon2023pagedattention}
reduce per-pass cost and compose with ours.

\section{Conclusion}

Speculative decoding no longer requires a discrete vocabulary.
By rebuilding acceptance, correction, and verification for continuous
patches, we turn the horizon-long sequential bottleneck of TSFMs into a
parallelism opportunity, delivering up to $3.0\times$ speedup across
five model families with provably bounded deviation from target-only
forecasts and predictable speedups. It is training-free, composes with
kernel-level optimizations, and admits acceptance-trained drafts.

\bibliography{references}

\clearpage
\def\maininclude{1}
\ifx\maininclude\undefined
\documentclass[letterpaper]{article} 
\usepackage[preprint]{aaai2027}
\usepackage[hyphens]{url}  
\usepackage{graphicx} 
\urlstyle{rm} 
\def\UrlFont{\rm}  
\usepackage{natbib}  
\usepackage{caption} 
\frenchspacing  

\usepackage{algorithm}
\usepackage{algorithmic}

\usepackage{amsmath}
\usepackage{amssymb}
\usepackage{amsthm}
\usepackage{booktabs}
\usepackage{multirow}

\usepackage{xr}
\makeatletter
\newcommand*{\addFileDependency}[1]{%
  \typeout{(#1)}%
  \@addtofilelist{#1}%
  \IfFileExists{#1}{}{\typeout{No file #1.}}%
}
\newcommand*{\myexternaldocument}[1]{%
  \externaldocument{#1}%
  \addFileDependency{#1.tex}%
  \addFileDependency{#1.aux}%
}
\let\XRsavedbibcite\bibcite
\let\bibcite\@gobbletwo
\myexternaldocument{main}
\let\bibcite\XRsavedbibcite
\makeatother

\setcounter{topnumber}{3}
\setcounter{dbltopnumber}{3}
\setcounter{totalnumber}{5}
\renewcommand{\topfraction}{0.95}
\renewcommand{\dbltopfraction}{0.95}
\renewcommand{\textfraction}{0.05}
\renewcommand{\floatpagefraction}{0.9}
\renewcommand{\dblfloatpagefraction}{0.9}

\pdfinfo{
/TemplateVersion (2027.1)
}

\setcounter{secnumdepth}{2}

\newcommand{\E}{\mathbb{E}}
\newcommand{\R}{\mathbb{R}}
\newcommand{\KL}{\mathrm{KL}}
\newcommand{\TV}{\mathrm{TV}}
\newcommand{\norm}[1]{\left\lVert #1 \right\rVert}
\newcommand{\Swall}{S}

\theoremstyle{plain}
\newtheorem{theorem}{Theorem}[section]
\newtheorem{lemma}[theorem]{Lemma}
\newtheorem{proposition}[theorem]{Proposition}
\newtheorem{corollary}[theorem]{Corollary}
\theoremstyle{definition}
\newtheorem{definition}[theorem]{Definition}
\newtheorem{assumption}[theorem]{Assumption}
\newtheorem{remark}[theorem]{Remark}

\numberwithin{equation}{section}
\numberwithin{figure}{section}
\numberwithin{table}{section}

\title{Accelerating Time Series Foundation Models with Speculative
Decoding\\[0.4em]\large Technical Appendix}
\author{
Pranav Subbaraman$^{*1}$\quad
Fang Sun$^{*1}$\quad
Jinxi Yu$^{*1}$\quad
Yue Yao$^{1}$\quad
Huacong Tang$^{1}$\quad
Xiao Luo$^{2}$\quad
Yizhou Sun$^{1}$
}
\affiliations{
$^{*}$Equal contribution, alphabetical order.\\
$^{1}$University of California, Los Angeles\quad
$^{2}$University of Wisconsin--Madison
}

\begin{document}

\maketitle

\noindent
This document is the technical appendix of the submission
\emph{Accelerating Time Series Foundation Models with Speculative
Decoding}. Appendix~\ref{app:proofs} contains all proofs and supporting
theoretical results; Appendix~\ref{app:details} contains the full
experimental record; Appendix~\ref{app:related} provides extended
related-work discussion. Numbers of the form A.1 or B.1 refer to this
document; unprefixed equation, table, figure, and result numbers refer
to the main paper. The reproducibility checklist is submitted as a
separate document.

\fi

\ifx\maininclude\undefined\else
\setcounter{secnumdepth}{2}
\numberwithin{theorem}{section}
\numberwithin{equation}{section}
\numberwithin{figure}{section}
\numberwithin{table}{section}
\makeatletter
\let\c@lemma\c@theorem      \let\thelemma\thetheorem
\let\c@proposition\c@theorem \let\theproposition\thetheorem
\let\c@corollary\c@theorem   \let\thecorollary\thetheorem
\let\c@definition\c@theorem  \let\thedefinition\thetheorem
\let\c@assumption\c@theorem  \let\theassumption\thetheorem
\let\c@remark\c@theorem      \let\theremark\thetheorem
\makeatother
\fi

\appendix

\section{Proofs and Derivations}
\label{app:proofs}

This appendix collects full proofs for the Theoretical Guarantees
subsection of the main paper, adapting the
discrete-token arguments of speculative decoding
\citep{leviathan2023fast,chen2023speculative} to densities on $\R^d$.
Throughout, $p$ and $q$ denote the target and draft surrogate densities of
\eqref{eq:heads} at a fixed position (position subscripts $p_i,q_i$ are
used where the position matters), $K$ is the block size, $c$ the
draft-to-target per-call GPU cost ratio, and $v$ the
verification-to-forward cost ratio, as in the main text. All
distributional statements are statements about the surrogate chain.

\subsection{Setup}

\begin{assumption}[Regularity]
\label{ass:reg}
At every position, the target and draft conditionals $p(\cdot\mid h)$ and
$q(\cdot\mid h)$ are Lebesgue densities on $(\R^d,\mathcal{B})$;
acceptance functions $\alpha(\cdot)\in[0,1]$ are measurable and are
evaluated only on $\{q>0\}$; conditional draws are independent given the
history.
\end{assumption}

\begin{definition}[Overlap and residual]
\label{def:overlap}
The overlap of $p$ and $q$ and the residual density are
\begin{align}
\beta &= \int_{\R^d}\min\{p(x),q(x)\}\,dx \notag\\
      &= 1-\tfrac12\int_{\R^d}|p(x)-q(x)|\,dx,\\
\rho(x) &= \frac{(p(x)-q(x))_+}{1-\beta},
\qquad (a)_+=\max\{a,0\}.
\end{align}
$\rho$ is well defined whenever $\beta<1$ (if $p=q$, speculative decoding
is vacuous), and it is a valid density:
$\int\rho=\int(p-q)_+/(1-\beta)=(1-\beta)/(1-\beta)=1$, using
$\int(p-q)_+=\tfrac12\int|p-q|=1-\beta$.
\end{definition}

For the likelihood-ratio rule $\alpha=\min\{1,p/q\}$, the mean acceptance
under $X\sim q$ equals the overlap:
$\bar\alpha=\int q\min\{1,p/q\}=\int\min\{p,q\}=\beta$. Values of $\alpha$
on $\{q=0\}$ are immaterial, since proposals never land there.

\subsection{Exactness of Residual Resampling}

The two theorems below record the classical exactness result for reference:
the residual-resampling construction (Algorithm~\ref{alg:lossless}) that our
scheme deliberately approximates rather than implements, for the reasons
given in the main text (Proposition~\ref{prop:rescost} quantifies the cost).

\begin{theorem}[Single-step exactness]
\label{thm:onestep}
Under Assumption~\ref{ass:reg}, sample $X\sim q$ and accept with
probability $\alpha(X)=\min\{1,p(X)/q(X)\}$; on rejection draw
$Y^\star\sim\rho$. Then the output ($X$ on acceptance, $Y^\star$ on
rejection) has density $p$.
\end{theorem}
\begin{proof}
Let $A$ denote acceptance. Then
$\Pr(X\in dx,\,A)=q(x)\min\{1,p(x)/q(x)\}\,dx=\min\{p(x),q(x)\}\,dx$, so
$\Pr(A)=\beta$. For any Borel set $B$,
\begin{align*}
\Pr(Y\in B)
&=\int_B\min\{p,q\}\,dx+\Pr(A^c)\int_B\rho(x)\,dx\\
&=\int_B\min\{p,q\}\,dx+(1-\beta)\int_B\tfrac{(p-q)_+}{1-\beta}\,dx\\
&=\int_B\bigl[\min\{p,q\}+(p-q)_+\bigr]\,dx\\
&=\int_B p(x)\,dx,
\end{align*}
using the pointwise identity $\min\{a,b\}+(a-b)_+=a$.
\end{proof}

\begin{theorem}[Block and autoregressive exactness]
\label{thm:block}
Consider a block of $K$ draft proposals $(X_1,\dots,X_K)$ drawn
autoregressively from $q$ and verified by $p$ on the corresponding
prefixes. Let $n$ be the number of consecutive accepts (possibly $n=K$).
If $n<K$, draw the $(n{+}1)$-st output from the residual $\rho_{n+1}$
formed from the pair $(p_{n+1},q_{n+1})$ at the failed position; if
$n=K$, draw one more patch from $p_{K+1}$. Iterating this procedure
yields the same joint law as pure target autoregression: the output
sequence has density $\prod_t p_t(\cdot\mid h_t)$.
\end{theorem}
\begin{proof}
By strong induction over positions. Fix a realized history $h_1$. For
position $i=1$, Theorem~\ref{thm:onestep} gives the marginal $p_1$.
Assume the prefix $(Y_1,\dots,Y_{i-1})$ has joint density
$\prod_{j<i}p_j(\cdot\mid h_j)$. At position $i$, acceptance is computed
from $(p_i,q_i)$ conditional on the realized prefix; by
Theorem~\ref{thm:onestep}, the conditional output law at $i$ (including
the first-failure case via $\rho_i$) equals $p_i$. Hence the joint law up
to $i$ is $\prod_{j\le i}p_j$. If no failure occurs inside the block, the
$(K{+}1)$-st draw is from $p_{K+1}$, the next conditional of the target
chain. Concatenating blocks yields $\prod_t p_t$.
\end{proof}

\begin{algorithm}[tb]
\caption{Residual-resampling round (analyzed; not deployed)}
\label{alg:lossless}
\textbf{Input}: history $h$; target $p$; draft $q$; block size $K$\\
\textbf{Output}: $1$ to $K{+}1$ committed patches
\begin{algorithmic}[1]
\STATE draft proposals $x_{1:K}\sim q$ autoregressively; one causal
target pass gives $p_1,\dots,p_{K+1}$ at the prefix boundaries
\FOR{$i=1,\dots,K$}
\STATE accept $x_i$ with probability $\min\{1,p_i(x_i)/q_i(x_i)\}$; stop
at the first rejection (index $n{+}1$)
\ENDFOR
\STATE \textbf{if} no rejection: draw $t\sim p_{K+1}$ and \textbf{return}
$[x_{1:K},t]$
\STATE \textbf{else}: draw $t\sim\rho_{n+1}$ formed from
$(p_{n+1},q_{n+1})$ and \textbf{return} $[x_{1:n},t]$
\end{algorithmic}
\end{algorithm}

\subsection{The Fallback Kernel and Its Deviation Bounds}

The next proposition, summarized in the Theoretical Guarantees
subsection of the main text, characterizes the one-step output law of
the fallback scheme and bounds its deviation from the target.

\begin{proposition}[Fallback kernel]
\label{prop:kernel}
With acceptance function $\alpha(\cdot)\in[0,1]$,
$\bar\alpha=\int\alpha\,q$, and fallback draw from $p$ on rejection, the
one-step output density is $g=\alpha q+(1-\bar\alpha)p$, and
\begin{gather}
\norm{g-p}_{\TV}=\tfrac12\!\int\!\bigl|\alpha q-\bar\alpha p\bigr|
\;\le\;\bar\alpha ,\\
\KL(g\,\|\,p)\le\!\int\!\alpha q\log\frac{\alpha q}{\bar\alpha p}.
\end{gather}
For the likelihood-ratio rule $\alpha=\min\{1,p/q\}$, moreover,
$\alpha q=\min\{p,q\}$ and $\norm{g-p}_{\TV}\le 1-\bar\alpha$: the
deviation vanishes in the high-acceptance regime, exactly where the
generic bound $\bar\alpha$ is loose.
The autoregressive chain composes these kernels, and the joint deviation
over $T$ steps is at most $\sum_{t\le T}\delta_t$ for per-step TV bounds
$\delta_t$, by stepwise coupling.
\end{proposition}

\begin{proof}
\emph{Output law.} For any Borel set $B$,
\begin{align*}
\Pr(Y\in B)
&=\Pr(\text{accept},\,Y\in B)+\Pr(\text{reject},\,Y\in B)\\
&=\int_B\alpha(x)q(x)\,dx+(1-\bar\alpha)\int_B p(x)\,dx,
\end{align*}
since rejection occurs with probability $1-\bar\alpha$ and the fallback
draw from $p$ is independent of the rejected proposal. Hence
$g=\alpha q+(1-\bar\alpha)p$.

\emph{TV bound.} $g-p=\alpha q-\bar\alpha p$, so
\[
\norm{g-p}_{\TV}
=\tfrac12\int|\alpha q-\bar\alpha p|
\le\tfrac12\Bigl(\int\alpha q+\bar\alpha\int p\Bigr)=\bar\alpha,
\]
using $\int\alpha q=\bar\alpha$ and $\int p=1$. For bounded $f$ with
$\norm{f}_\infty\le M$ this gives $|\E_g[f]-\E_p[f]|\le2M\bar\alpha$.

\emph{KL bound.} Decompose $p=\bar\alpha p+(1-\bar\alpha)p$ and apply the
log-sum inequality to the pairs $(\alpha q,\ \bar\alpha p)$ and
$\bigl((1-\bar\alpha)p,\ (1-\bar\alpha)p\bigr)$:
\begin{align*}
\KL(g\,\|\,p)
&=\int\bigl[\alpha q+(1-\bar\alpha)p\bigr]
  \log\frac{\alpha q+(1-\bar\alpha)p}{p}\\
&\le\int\alpha q\log\frac{\alpha q}{\bar\alpha p}
  +\int(1-\bar\alpha)p\,\log1\\
&=\int\alpha q\log\frac{\alpha q}{\bar\alpha p}.
\end{align*}
Pinsker's inequality additionally gives
$\norm{g-p}_{\TV}\le\sqrt{\tfrac12\KL(g\|p)}$.

\emph{Refinement for the likelihood-ratio rule.} If
$\alpha=\min\{1,p/q\}$ then $\alpha q=\min\{p,q\}$ and
$\bar\alpha=\beta$. Write
\[
g-p=\min\{p,q\}-\bar\alpha p
=\bigl(\min\{p,q\}-p\bigr)+(1-\bar\alpha)p.
\]
The first term has $L_1$ mass $\int(p-\min\{p,q\})=1-\beta=1-\bar\alpha$
and the second has mass $1-\bar\alpha$, so
$\norm{g-p}_{\TV}\le\tfrac12\bigl[(1-\bar\alpha)+(1-\bar\alpha)\bigr]
=1-\bar\alpha$.

\emph{Composition over the horizon.} Let
\[
\begin{aligned}
K_t(h_t,dy)=\bigl[&\alpha_t(y\mid h_t)q_t(y\mid h_t)\\
&+(1-\bar\alpha_t(h_t))p_t(y\mid h_t)\bigr]dy
\end{aligned}
\]
denote the one-step kernel, with
$\bar\alpha_t(h_t)=\int\alpha_t(y\mid h_t)q_t(y\mid h_t)\,dy$.
Conditioning on $h_t$ and applying the one-step law at each position
shows the generated sequence has joint density
$\prod_{t=1}^{T}K_t(h_t,dy_t)$. The horizon-wise bound is
Proposition~\ref{prop:horizon} below.
\end{proof}

\begin{proposition}[Horizon-wise deviation]
\label{prop:horizon}
Let
$\delta_t:=\sup_{h_t}\norm{K_t(\cdot\mid h_t)-p_t(\cdot\mid h_t)}_{\TV}$.
Then for any horizon $T$,
\[
\begin{aligned}
\bigl\|\mathrm{Law}_{\mathrm{SD}}(Y_{1:T})
-\mathrm{Law}_{p}(Y_{1:T})\bigr\|_{\TV}
&\le 1-\prod_{t=1}^{T}(1-\delta_t)\\
&\le\sum_{t=1}^{T}\delta_t.
\end{aligned}
\]
\end{proposition}
\begin{proof}
Couple the two chains stepwise: given agreement of the histories through
step $t-1$, a maximal coupling of $K_t(\cdot\mid h_t)$ and
$p_t(\cdot\mid h_t)$ makes step $t$ agree with probability at least
$1-\delta_t$. The chains then agree through $T$ steps with probability at
least $\prod_{t\le T}(1-\delta_t)$; taking complements and the union
bound gives both inequalities.
\end{proof}

\begin{proposition}[Continuity in the draft head]
\label{prop:continuity}
Under the likelihood-ratio rule,
$K_t(\cdot\mid h_t)=\min\{p_t,q_t\}+(1-\bar\alpha_t)p_t$ with
$\bar\alpha_t=\int\min\{p_t,q_t\}$. If
$\norm{q_t(\cdot\mid h_t)-p_t(\cdot\mid h_t)}_{L_1}\to0$, then
$\bar\alpha_t(h_t)\to1$ and
$\norm{K_t(\cdot\mid h_t)-p_t(\cdot\mid h_t)}_{\TV}\to0$: the fallback
kernel converges to the target kernel as the draft head converges to the
target head.
\end{proposition}
\begin{proof}
$\bar\alpha_t=1-\tfrac12\norm{p_t-q_t}_{L_1}\to1$, and by the
likelihood-ratio refinement in the proof of
Proposition~\ref{prop:kernel},
$\norm{K_t-p_t}_{\TV}\le1-\bar\alpha_t\to0$.
\end{proof}

\subsection{Reduced-Noise Fallback}

\begin{lemma}[Reduced-noise fallback]
\label{lem:fallback}
If rejection instead commits a draw from
$p_r=\mathcal{N}(\mu_p,r^2I_d)$ with $r\le\sigma_p$, the output law is
$g_r=\alpha q+(1-\bar\alpha)p_r$ and
\begin{equation}
\norm{g_r-p}_{\TV}\;\le\;\bar\alpha+(1-\bar\alpha)\,\norm{p_r-p}_{\TV}.
\end{equation}
$r=\sigma_p$ recovers Proposition~\ref{prop:kernel}; as $r\to0$ this TV
bound becomes vacuous: that endpoint is governed by squared-error fidelity
(Proposition~\ref{prop:detfidelity} of the main paper), not
distributional closeness.
\end{lemma}

\begin{proof}
The same conditioning as in the proof of Proposition~\ref{prop:kernel}
gives $g_r=\alpha q+(1-\bar\alpha)p_r$, so
$g_r=g+(1-\bar\alpha)(p_r-p)$ and, by the triangle inequality,
\[
\begin{aligned}
\norm{g_r-p}_{\TV}
&\le\norm{g-p}_{\TV}+(1-\bar\alpha)\norm{p_r-p}_{\TV}\\
&\le\bar\alpha+(1-\bar\alpha)\norm{p_r-p}_{\TV}.
\end{aligned}
\]
For $r=\sigma_p$, $p_r=p$ and Proposition~\ref{prop:kernel} is recovered.
As $r\to0$, $p_r$ concentrates its mass on balls of radius $O(r)$ around
$\mu_p$, whose $p$-mass vanishes; hence $\norm{p_r-p}_{\TV}\to1$ and the
bound degenerates to $\bar\alpha+(1-\bar\alpha)=1$, which is vacuous for
a TV distance.
\end{proof}

\subsection{Proof of Proposition~\ref{prop:detfidelity} of the Main
Paper (Deterministic Fidelity)}

\begin{proof}
Under the deterministic gate \eqref{eq:detgate} the proposal $\mu_q$ is
accepted with probability $\alpha=e^{-\Delta^2/2\sigma^2}$, in which case the
committed patch is $Y=\mu_q$; otherwise ($r=0$) the fallback commits
$Y=\mu_p$. Hence $\norm{Y-\mu_p}_d^2$ equals $\Delta^2$ with probability
$\alpha$ and $0$ otherwise, so
$\E[\norm{Y-\mu_p}_d^2\mid h]=\alpha\,\Delta^2=\Delta^2 e^{-\Delta^2/2\sigma^2}$.
Writing $u=\Delta^2\ge0$ and $f(u)=u\,e^{-u/2\sigma^2}$, the derivative
$f'(u)=e^{-u/2\sigma^2}\bigl(1-u/2\sigma^2\bigr)$ vanishes only at
$u=2\sigma^2$, the global maximum of $f$ on $[0,\infty)$, with value
$f(2\sigma^2)=2\sigma^2/e$; the bound is therefore independent of $\Delta$.
For the tail, the event $\{\norm{Y-\mu_p}_d^2>\tau\}$ requires both
acceptance and $\Delta^2>\tau$, so its probability is
$\mathbf 1\{\Delta^2>\tau\}\,e^{-\Delta^2/2\sigma^2}\le e^{-\tau/2\sigma^2}$.
\end{proof}

\begin{remark}[Why means, not samples]
\label{rem:detpoint}
Our scheme commits either $\mu_q$ or $\mu_p$, the risk-optimal (Bayes)
actions under squared error; injecting noise only inflates squared error,
as the noise ablation in the main text confirms empirically.
\end{remark}

\subsection{Equal-Covariance Acceptance}

\begin{proposition}[Equal-covariance acceptance]
\label{prop:closedform}
For $p=\mathcal{N}(\mu_p,\Sigma)$ and $q=\mathcal{N}(\mu_q,\Sigma)$ the mean
acceptance of the stochastic rule has the closed form
$\bar\alpha=2\Phi(-\Delta/2)$ with
$\Delta^2=(\mu_p-\mu_q)^\top\Sigma^{-1}(\mu_p-\mu_q)$; for the deterministic
proposal, acceptance is the deterministic gate \eqref{eq:detgate}. The
plug-in estimate of $\bar\alpha$ from held-out histories is unbiased and
concentrates at Hoeffding rate (Proposition~\ref{prop:est},
Theorem~\ref{thm:hoeffding}), making
\eqref{eq:EL}--\eqref{eq:speedup} predictable before deployment.
\end{proposition}

\begin{proof}
Whitening by $\Sigma^{-1/2}$ maps $p$ and $q$ to $\mathcal{N}(0,I_d)$ and
$\mathcal{N}(\delta,I_d)$ with $\norm{\delta}_2=\Delta$; the overlap is
invariant under this affine map. Under $p$, the statistic
$t=\delta^\top x/\Delta$ is standard normal, and $p(x)\le q(x)$ iff
$\norm{x}^2\ge\norm{x-\delta}^2$ iff $t\ge\Delta/2$; under $q$,
$t\sim\mathcal{N}(\Delta,1)$. Hence
\begin{align*}
\bar\alpha
&=\int\min\{p,q\}\\
&=\Pr\nolimits_p(t\ge\Delta/2)+\Pr\nolimits_q(t<\Delta/2)\\
&=\Phi(-\Delta/2)+\Phi(\Delta/2-\Delta)\\
&=2\Phi(-\Delta/2).
\end{align*}
For the deterministic proposal $x=\mu_q$, substituting into
\eqref{eq:logratio} with $\sigma_p=\sigma_q=\sigma$ cancels the second
and third terms and leaves the gate \eqref{eq:detgate}. The concentration
claim is Theorem~\ref{thm:hoeffding} below.
\end{proof}

\subsection{Block Length and Block-Size Selection}

\begin{proposition}[Capped geometric block length]
\label{prop:blocklength}
Suppose $\Pr(A_i\mid A_1,\dots,A_{i-1})=\bar\alpha$ within a round. Then
$\Pr(L=\ell)=(1-\bar\alpha)\bar\alpha^{\ell-1}$ for $1\le\ell\le K$,
$\Pr(L=K{+}1)=\bar\alpha^{K}$, and \eqref{eq:EL} holds.
\end{proposition}
\begin{proof}
The round emits $\ell\le K$ patches iff the first $\ell-1$ positions are
accepted and position $\ell$ is rejected (the rejected position is
replaced by the fallback commit), and $K{+}1$ patches iff all $K$
positions are accepted (the bonus patch). Hence
$\Pr(L\ge\ell)=\bar\alpha^{\ell-1}$ for $1\le\ell\le K{+}1$, and the
tail-sum formula gives
\[
\E[L]=\sum_{\ell=1}^{K+1}\Pr(L\ge\ell)
=\sum_{j=0}^{K}\bar\alpha^{j}
=\frac{1-\bar\alpha^{K+1}}{1-\bar\alpha}.
\qedhere
\]
\end{proof}

\begin{proposition}[Dependence bounds]
\label{prop:dependence}
If
$\underline{\alpha}\le\Pr(A_i\mid A_1,\dots,A_{i-1})\le\overline{\alpha}$
for $i=1,\dots,K$, then
\[
\frac{1-\underline{\alpha}^{\,K+1}}{1-\underline{\alpha}}
\;\le\;\E[L]\;\le\;
\frac{1-\overline{\alpha}^{\,K+1}}{1-\overline{\alpha}}.
\]
\end{proposition}
\begin{proof}
$\Pr(L\ge\ell)=\prod_{i=1}^{\ell-1}\Pr(A_i\mid A_1,\dots,A_{i-1})$, so
$\underline{\alpha}^{\ell-1}\le\Pr(L\ge\ell)\le\overline{\alpha}^{\ell-1}$
for every $\ell\le K{+}1$; summing the tail probabilities termwise gives
the claim.
\end{proof}

Acceptance events within a block are correlated in practice;
Proposition~\ref{prop:dependence} shows that $\E[L]$ is nevertheless
bounded by the same functional form at the extreme conditional rates, so
plug-in throughput predictions remain reliable when the conditional
acceptance stays in a narrow band around $\bar\alpha$.

\begin{proposition}[Marginal-gain condition for $K$]
\label{prop:kstar}
$\Swall(K{+}1)\ge\Swall(K)$ if and only if
$\bar\alpha^{K+1}(cK+v)\ge c\,\E[L_K]$, where $\E[L_K]$ denotes
\eqref{eq:EL} at block size $K$. A near-optimal block size $K^\star$ is
the largest $K$ for which the condition holds with the measured pair
$(\hat{\bar\alpha},c)$.
\end{proposition}
\begin{proof}
Since $\E[L_{K+1}]=\E[L_K]+\bar\alpha^{K+1}$,
\begin{align*}
\Swall(K{+}1)\ge\Swall(K)
&\iff\frac{\E[L_K]+\bar\alpha^{K+1}}{c(K{+}1)+v}
  \ge\frac{\E[L_K]}{cK+v}\\
&\iff\bigl(\E[L_K]+\bar\alpha^{K+1}\bigr)(cK+v)\\
&\qquad\ge\E[L_K](cK+v)+c\,\E[L_K]\\
&\iff\bar\alpha^{K+1}(cK+v)\ge c\,\E[L_K].
\qedhere
\end{align*}
\end{proof}

Interpretation: the extra position pays off when the marginal patch
$\bar\alpha^{K+1}$ it contributes exceeds its marginal draft cost $c$
times the current yield-per-cost $\E[L_K]/(cK+v)$.

\begin{remark}[Compute overhead]
\label{rem:ops}
A round spends $cK+v$ target-forward equivalents of GPU compute and
commits $\E[L]$ patches, so the compute per committed patch relative to
target-only decoding is $(1-\bar\alpha)(cK+v)/(1-\bar\alpha^{K+1})$:
speculative decoding trades extra total compute for fewer sequential
target passes.
\end{remark}

\begin{remark}[Estimating $(\bar\alpha,c)$]
\label{rem:estimating}
Exact evaluation of $\bar\alpha$ requires a $d$-dimensional integral with
no closed form beyond Proposition~\ref{prop:closedform}, and an
analytical $c$ (e.g.\ a FLOPs ratio) ignores kernel scheduling and memory
effects. Both are therefore estimated empirically: $\bar\alpha$ as the
accepted fraction on held-out batches (Theorem~\ref{thm:hoeffding}) and
$c$ from GPU-compute timing of draft and target forwards.
\end{remark}

\subsection{Acceptance Estimation and Concentration}

For a fixed history $h$, the mean acceptance of the stochastic rule
equals the overlap
$\beta(h)=\int\min\{p(y\mid h),q(y\mid h)\}\,dy$
(Definition~\ref{def:overlap}); the deployment mean acceptance is
$\bar\alpha:=\E_h[\beta(h)]\in[0,1]$.

\begin{proposition}[Unbiased plug-in estimator]
\label{prop:est}
Given i.i.d.\ histories $h_1,\dots,h_N$, the estimator
$\hat{\bar\alpha}_N=\frac1N\sum_{i=1}^{N}\beta(h_i)$ satisfies
$\E[\hat{\bar\alpha}_N]=\bar\alpha$. When $\beta(h_i)$ is unavailable in
closed form, an inner Monte Carlo estimate (drawing
$X_{ij}\sim q(\cdot\mid h_i)$ and averaging
$\min\{1,p(X_{ij})/q(X_{ij})\}$) is unbiased for $\beta(h_i)$ and
preserves the unbiasedness of $\hat{\bar\alpha}_N$.
\end{proposition}
\begin{proof}
Linearity of expectation, and, for the inner estimate,
$\E_{X\sim q}[\min\{1,p/q\}]=\int\min\{p,q\}=\beta(h_i)$.
\end{proof}

\begin{theorem}[Hoeffding concentration]
\label{thm:hoeffding}
Each per-history term (the exact $\beta(h_i)$ or its inner Monte Carlo
average) lies in $[0,1]$ and is i.i.d.\ across $i$, so for any
$\varepsilon>0$,
\[
\Pr\bigl(|\hat{\bar\alpha}_N-\bar\alpha|\ge\varepsilon\bigr)
\le2\exp(-2N\varepsilon^{2}),
\]
and $\hat{\bar\alpha}_N\to\bar\alpha$ almost surely.
\end{theorem}
\begin{proof}
Hoeffding's inequality for i.i.d.\ $[0,1]$-valued variables, and the
strong law of large numbers.
\end{proof}

\begin{corollary}[Consistency of throughput predictors]
\label{cor:plugin}
The plug-in predictors obtained by substituting $\hat{\bar\alpha}_N$ into
\eqref{eq:EL} and \eqref{eq:speedup} are consistent as $N\to\infty$, by
the continuous mapping theorem.
\end{corollary}

\subsection{Cost of Residual Resampling}

\begin{proposition}[Expected residual cost]
\label{prop:rescost}
Draw $Z\sim p$ and accept with probability $\pi(Z)=(1-q(Z)/p(Z))_+$; the
accepted draw has density $\rho$, and the expected number of target draws
per residual sample is $1/(1-\beta)$.
\end{proposition}
\begin{proof}
$\Pr(Z\in dz,\ \text{accept})=p(z)\,(1-q(z)/p(z))_+\,dz=(p(z)-q(z))_+\,dz$,
so the acceptance probability is $\int(p-q)_+=1-\beta$ and the accepted
draw has density $(p-q)_+/(1-\beta)=\rho$. The number of attempts is
geometric with success probability $1-\beta$, with mean $1/(1-\beta)$.
\end{proof}

\begin{remark}[Why the fallback scheme]
\label{rem:whyfallback}
Classical rejection sampling from $q$ with an envelope constant
$M\ge\sup_x p(x)/q(x)$ is impractical here: a finite, tight $M$ is
unknown in general and can be infinite. Thinning by
Proposition~\ref{prop:rescost} costs $1/(1-\beta)$ target draws at the
first failed position of a round; at the overlaps $\beta\in[0.7,0.95]$ of
our operating points this is a factor of $3$--$20$, unbounded as the
models agree, so residual exactness can cost more than parallel
verification saves. A crude breakeven heuristic is that residual
resampling is competitive only when $1-\bar\alpha\gtrsim1/K$, so that the
expected residual cost per round does not exceed the round's expected
yield. This motivates the fallback scheme analyzed in
Proposition~\ref{prop:kernel}.
\end{remark}

\subsection{Additional Remarks}

\begin{remark}[$1/d$ scaling of the acceptance statistic]
\label{rem:dscaling}
The $1/d$ scaling in \eqref{eq:logratio} is applied to the energies and
the log-determinant term alike, so the two contributions remain
commensurate. Without it, the raw $d$-dimensional log ratio grows linearly
in $d$ and the acceptance probability degenerates to a nearly $0/1$ gate
at the patch dimensions used here; the scaled statistic instead uses the
tempered ratio $(p/q)^{1/d}$, so a single $O(1)$ temperature $\sigma$ is
meaningful across families. Statements tied to the exact ratio (the
overlap identity and Proposition~\ref{prop:closedform}) are made for the
unscaled rule; empirical acceptance is always reported alongside.
\end{remark}

\begin{remark}[The normalization term is load-bearing]
\label{rem:normterm}
With a learned-variance draft ($\sigma_q\ll\sigma_p$) and the determinant
term of \eqref{eq:logratio} omitted, the tempered statistic $\ell_i$ is
inflated by $\log(\sigma_p/\sigma_q)>0$ (a factor $d$ larger in the
untempered log ratio) and $\alpha\to1$ regardless of
draft--target agreement: speculative decoding silently becomes draft-only
decoding.
\end{remark}

\begin{remark}[General covariances]
\label{rem:gencov}
For diagonal or full covariances, the acceptance statistic
\eqref{eq:logratio} generalizes by replacing the scaled Euclidean
energies with scaled Mahalanobis energies
$\frac1d(x-\mu)^\top\Sigma^{-1}(x-\mu)$ and the normalization term with
$\frac{1}{2d}\log\bigl(|\Sigma_q|/|\Sigma_p|\bigr)$; the isotropic case
recovers \eqref{eq:logratio}. More expressive covariances can raise
$\bar\alpha$ by matching the target more closely, at higher per-step
evaluation cost; the headline runs use the isotropic surrogate of
\eqref{eq:heads}. Omitting the log-determinant term under unequal
covariances produces the failure mode of Remark~\ref{rem:normterm}.
\end{remark}

\begin{remark}[Verification cost accounting]
\label{rem:verifycost}
For families exposing per-position boundary predictions, one verification
round is a single causal forward over $[h,x_1,\dots,x_K]$: relative to a
baseline forward over $h$ it adds only the $K$ proposal tokens, so
$v\approx1$ and the target-call count per round is one, independent of
$K$. For TiRex-style batched $K$-prefix verification, the round is still
one call but its compute grows with $K$; this enters \eqref{eq:speedup}
through $v$ and explains the TiRex failure cases in the main text.
\end{remark}

\section{Experiment Details}
\label{app:details}

\begin{figure}[tb]
\centering
\includegraphics[width=0.95\columnwidth]{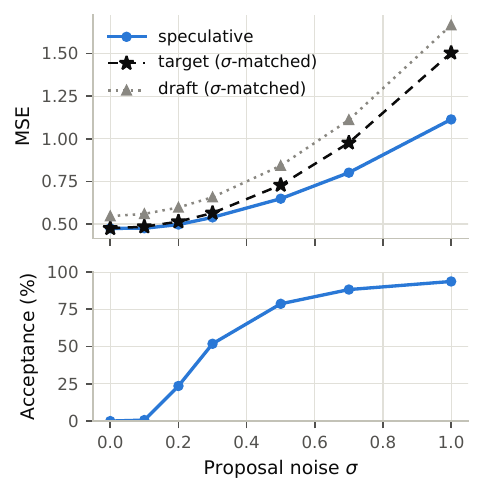}
\caption{Noise ablation on Timer-XL ETTh1 (batch 64, $K{=}1$). The top
panel shows MSE of speculative decoding with stochastic proposals and
deterministic fallback against noise-matched target-only and draft-only
decoding, and the bottom panel the empirical acceptance.}
\label{fig:noise}
\end{figure}

This appendix collects the full experimental record behind the main
text: the architectural compatibility matrix and decoding pseudocode,
the dataset descriptions, the accuracy-gate exclusion list, the
noise-ablation protocol
(Figure~\ref{fig:noise}), the complete block-size and batch-size sweeps
(Tables~\ref{tab:ksweep} and~\ref{tab:bsweep}), per-family
configurations (Tables~\ref{tab:hp-timerxl}--\ref{tab:hp-tirex}), and
the draft training recipe (Table~\ref{tab:trainhparams}). Settings
shared by all five families: mean-normalized ($1/d$-scaled) acceptance
energy, acceptance bias disabled, free bonus patch on full acceptance,
GPU compute time from CUDA events, seed $2021$ (the deterministic TiRex
pipeline uses none), single H200 GPU.
Timer-XL, TimesFM, Sundial, and Time-MoE run through one shared
speculative decoder; TiRex uses a standalone implementation.

\paragraph{Computing Infrastructure.}
Every reported run executes on a single NVIDIA H200 GPU ($140$\,GB) of
an x86-64 Linux $5.15$ host, under Python $3.10.20$, PyTorch $2.3.1$
with CUDA $12.1$, NumPy $1.26.4$, pandas $2.1.4$, einops $0.7.0$, and
transformers $4.44.2$. Because timing brackets model forwards with CUDA
events and excludes Python-side logic and data loading, host CPU and
memory do not enter any reported number.

\paragraph{Metrics and Number of Runs.}
For forecasts $\hat y$ and ground truth $y$ over the evaluated windows,
channels, and horizon steps,
$\mathrm{MSE}=\operatorname{mean}\bigl[(\hat y-y)^2\bigr]$, computed on
the standardized series produced by the benchmark loader, with lower
values better. We report MSE because the deployed scheme commits
conditional means, the Bayes action under squared loss
(Proposition~\ref{prop:detfidelity}). Every accuracy, acceptance,
and timing number in this paper comes from one evaluation run at seed
$2021$; none is averaged over repeated runs. Within a run, reported GPU
time is the mean over the measured batches, taken after the per-family
warm-up batches of
Tables~\ref{tab:hp-timerxl}--\ref{tab:hp-tirex}.

\subsection{Architectural Compatibility and Decoding Algorithm}
\label{app:implementation}

\begin{table*}[t]
\centering
\small
\setlength{\tabcolsep}{2.5pt}
\begin{tabular}{@{}llllll@{}}
\toprule
Family & Backbone / head & Causal AR & Single-pass verify & Draft (reused / lightly trained) & SD \\
\midrule
Timer-XL & decoder Transformer / point & yes & yes & distilled/supervised $0.125\times$ (158K) & yes \\
TimesFM 2.5 & decoder Transformer / quantiles & yes & yes & same ckpt, context $256$, flip off & yes \\
Sundial & decoder Transformer / flow head & yes & yes (channel-batched) & same ckpt, $1$ sample, context $512$ & yes \\
Time-MoE & decoder MoE / multi-res point & yes & yes (horizon head) & Time-MoE-50M, context $768$ & yes \\
TiRex & xLSTM / quantiles & yes & batched $K$-prefix call & same ckpt, coarse rollout $m{=}16$ & partial \\
\midrule
Chronos-Bolt & direct multi-horizon & no & --- & --- & no \\
Moirai & masked encoder, sampled & no & --- & --- & no \\
Chronos-T5 & causal AR, discrete tokens & yes & teacher-forced & --- & no (rule mismatch) \\
\bottomrule
\end{tabular}
\caption{Architectural support for Gaussian speculative decoding across
time series foundation model families. ``Partial'' for TiRex means that
verification is one call per round but its compute grows with $K$.}
\label{tab:families}
\end{table*}

\begin{algorithm}[tb]
\caption{One speculative round (single example)}
\label{alg:sd}
\textbf{Input}: history $h$; target $p$; draft $q$; block size $K$;
acceptance scale $\sigma$; fallback noise $r$\\
\textbf{Output}: $1$ to $K{+}1$ committed patches
\begin{algorithmic}[1]
\FOR{$i=1,\dots,K$}
\STATE $\mu_q^{(i)}\leftarrow$ draft prediction on $h\cup\{x_{1:i-1}\}$;
\ $x_i\leftarrow \mu_q^{(i)}$ (det.) or $\mu_q^{(i)}+\sigma_q\varepsilon$
(stoch.)
\ENDFOR
\STATE $\{\mu_p^{(1)},\dots,\mu_p^{(K+1)}\}\leftarrow$ one causal target pass
on $[h,x_1,\dots,x_K]$, read at prefix boundaries
\FOR{$i=1,\dots,K$}
\STATE $\alpha_i\leftarrow\min\{1,e^{\ell_i}\}$ with $\ell_i$ from
\eqref{eq:logratio}; draw $u\sim\mathcal{U}[0,1]$
\IF{$u\le\alpha_i$}
\STATE commit $x_i$
\ELSE
\STATE commit $t=\mu_p^{(i)}+r\,\varepsilon$ and \textbf{return}
\ENDIF
\ENDFOR
\STATE commit bonus $t=\mu_p^{(K+1)}+r\,\varepsilon$ \ (free; same pass)
\end{algorithmic}
\end{algorithm}

\paragraph{Datasets.}
ETTh1 and ETTh2 are hourly and ETTm1 and ETTm2 are sampled at $15$
minutes, with $7$ channels each; Weather has $21$ meteorological
channels at $10$ minutes; ECL is the hourly electricity consumption of
$321$ clients; Traffic the hourly occupancy of $862$ road sensors;
Solar the $10$-minute output of $137$ photovoltaic plants; and
PEMS03/04/07/08 traffic flow from $358$/$307$/$883$/$170$ sensors.

\paragraph{Per-Family Boundary Indexing.}
The boundary of prefix $i$ sits at token index
$N-1+i\cdot(P/P_{\text{in}})$ for input patch length $P_{\text{in}}$ and
context length $N$ tokens. Time-MoE reads its multi-resolution horizon
head at every boundary in one forward; Sundial folds channels into the
batch and samples its flow head once for all $K{+}1$ boundaries; TimesFM
reads the median head at patch boundaries of its internal $32$-step
tokenization.

\paragraph{Accuracy-Gate Exclusions.}
The $12$ family--dataset pairs where the draft matched or beat the
target's MSE, excluded from speculative runs by protocol: Timer-XL ETTh2,
ETTm1, ETTm2, Solar; TimesFM ETTm1; Sundial Weather; Time-MoE
ETTh1, ETTh2; TiRex ETTm1, ETTm2, Weather; and the $0.25\times$ Timer-XL
draft on Weather.

\paragraph{Noise-Ablation Protocol.}
The noise ablation of the main text runs four arms on Timer-XL ETTh1 (full
split, batch $64$, $K{=}1$, shared baseline): (i) sampled proposals with
noise-free fallback ($r{=}0$), sweeping the proposal noise $\sigma$;
(ii) mean proposals with fallback noise, sweeping $r$ at fixed
$\sigma{=}0.20$; and noise-matched (iii) target-only and (iv) draft-only
decoding, which replay the baselines with the same injected noise. In
arm~(i) the swept $\sigma$ is injected proposal noise, a knob distinct
from our scheme's noise-free acceptance temperature despite the shared
symbol. Figure~\ref{fig:noise} plots arms (i), (iii), and (iv).

\paragraph{Full Sweep Tables.}
Table~\ref{tab:ksweep} reports the complete block-size sweeps
(acceptance, MSE, and speedup at $K{=}1$--$4$ for Timer-XL, TimesFM,
and Time-MoE), and Table~\ref{tab:bsweep} the batch-size sweeps
(speedup at $B{=}1$--$128$ for four families).

\subsection{Sensitivity Analyses}
\label{app:sensitivity}

The operating points of Table~\ref{tab:main} fix four deployment knobs,
and we vary each around the headline settings to map how the speedup
responds.

\paragraph{Block Size.}
Table~\ref{tab:ksweep} sweeps $K{=}1$--$4$ at fixed acceptance
temperature. $K{=}1$ is optimal for TimesFM, since at roughly $70\%$
acceptance extra proposals mostly add draft cost and speedup falls
monotonically in $K$. Time-MoE inverts the pattern, as its acceptance is
high and its horizon head makes longer blocks cheap to verify, so
speedup still grows at $K{=}4$. Timer-XL on ECL is nearly flat, with a
shallow optimum at $K{=}3$. These orderings match what the marginal
criterion of Proposition~\ref{prop:kstar} predicts from each family's
measured $(\bar\alpha,c)$.

\paragraph{Batch Size.}
Table~\ref{tab:bsweep} sweeps the batch size at fixed $K$ and $\sigma$.
Speedup degrades once the batch alone saturates the GPU, because parallel
verification stops being free and per-example first-rejection stalls
desynchronize the batch. TiRex shows the strongest decay, compounded by
its $K$-scaled verification. TimesFM instead rises from batch $1$, where
draft context-switching overhead dominates, to a plateau, and Time-MoE
continues to gain up to batch $64$ and holds there at $128$.

\paragraph{Draft Configuration.}
For TimesFM, draft context $256$ maximizes speedup while $128$ gives the
best speculative MSE. For Timer-XL, widening the draft to $0.25\times$
makes it more accurate than the target on Weather, which removes the
reason to verify and triggers the accuracy gate. Sundial's flow-sample
count acts as a target-quality dial, with targets at $n_s{=}5/10/20$
yielding speculative speedups of $1.08/1.19/1.83\times$, so speculation
pays most for the highest-quality, slowest target.

\paragraph{Horizon.}
Longer horizons amortize per-round overheads over more rounds while the
number of sequential passes keeps growing, so speculation strengthens
exactly on the long-horizon workloads that motivated it. Moving from
$H{=}336$ to $H{=}720$ raises the speedup of every Timer-XL pair
measured at matched $K$ and batch size by $0.26\times$--$0.29\times$ at
comparable accuracy (Table~\ref{tab:main}).

\subsection{Qualitative Case Study}
\label{app:case-study}

Beyond aggregate metrics, Figure~\ref{fig:case} shows how the scheme
behaves on concrete series, using Time-MoE on Weather at the block size
and acceptance temperature of Table~\ref{tab:ksweep} ($K{=}3$,
$\sigma{=}0.15$, horizon $1536$) on a bounded set of $64$ windows at
$B{=}8$. The run reproduces that table's acceptance to within
$0.1$ point ($91.8\%$ here against $91.9\%$ at $B{=}128$), so the
qualitative trace can be read against the sweeps; the smaller window set
and batch make its absolute errors ($0.4527$ speculative, $0.4236$ target,
$0.4995$ draft) their own reference point rather than entries comparable
with Tables~\ref{tab:main} and~\ref{tab:ksweep}. Two observations hold
across the window set.

\emph{Acceptance rises along the horizon.} Rejections concentrate in the
first few patches, and acceptance climbs monotonically with horizon
position, from $62\%$ on the first patch to $97\%$ by the eighth and
$100\%$ from the twelfth onward. Early patches are where the two models
still carry distinct information from the observed context; deeper into
the horizon both settle onto the same periodic continuation and the
proposals agree. This is the mechanism behind the block-size behaviour of
Table~\ref{tab:ksweep}: speculation grows cheaper the further it runs, so
long horizons favour larger $K$.

\emph{The fallback re-anchors the trajectory.} Rejection commits the
target's own prediction, and the trace shows the effect that is intended:
per-patch deviation from the target-only rollout is $0.015$ on rejected
patches against $0.028$ on accepted ones, and the patch immediately
following a rejection sits closer to target-only ($0.018$) than one
following an acceptance ($0.029$). The accuracy ordering survives at the
window level rather than only in aggregate, with speculative decoding more
accurate than draft-only on all $64$ windows and between target and draft
on $92\%$ of them.

\paragraph{Per-Family Configurations.}
Tables~\ref{tab:hp-timerxl}--\ref{tab:hp-tirex} list, per model family,
the exact target and draft configurations, speculative decoding
settings, and evaluation protocol behind every reported run.
Table~\ref{tab:trainhparams} gives the training hyperparameters of the
Timer-XL draft, the only trained component in this paper; the other four
families reuse public checkpoints without any training.

\begin{table}[tb]
\centering
\small
\setlength{\tabcolsep}{5pt}
\begin{tabular}{@{}crrrr@{}}
\toprule
$K$ & 1 & 2 & 3 & 4 \\
\midrule
\multicolumn{5}{@{}l}{\emph{Timer-XL ECL ($B{=}16$, $\sigma{=}0.25$)}} \\
Accept & 68.7\% & 71.3\% & 70.6\% & 73.2\% \\
MSE & 0.193 & 0.201 & 0.204 & 0.206 \\
Speedup & 1.29$\times$ & 1.32$\times$ & 1.34$\times$ & 1.32$\times$ \\
\midrule
\multicolumn{5}{@{}l}{\emph{TimesFM ETTh1 ($B{=}8$, $\sigma{=}0.10$)}} \\
Accept & 71.6\% & 75.6\% & 77.5\% & 78.2\% \\
MSE & 0.493 & 0.494 & 0.496 & 0.494 \\
Speedup & 1.21$\times$ & 1.18$\times$ & 1.16$\times$ & 1.04$\times$ \\
\midrule
\multicolumn{5}{@{}l}{\emph{Time-MoE Weather ($B{=}128$, $\sigma{=}0.15$)}} \\
Accept & 91.3\% & 91.7\% & 91.9\% & 92.1\% \\
MSE & 0.393 & 0.400 & 0.406 & 0.409 \\
Speedup & 1.72$\times$ & 2.34$\times$ & 2.61$\times$ & 2.84$\times$ \\
\bottomrule
\end{tabular}
\caption{Block-size ($K$) sweeps at fixed acceptance temperature.
$K{=}1$ is optimal for TimesFM, while Time-MoE still gains at $K{=}4$.
Each sweep is one internally consistent run, not comparable with
Table~\ref{tab:main} of the main paper.}
\label{tab:ksweep}
\end{table}

\begin{table}[tb]
\centering
\small
\setlength{\tabcolsep}{3.4pt}
\begin{tabular}{@{}lrrrrrrr@{}}
\toprule
$B$ & 1 & 4 & 8 & 16 & 32 & 64 & 128 \\
\midrule
Timer-XL ECL & 1.53 & 1.33 & --- & 1.34 & --- & --- & --- \\
TimesFM ETTh1 & 1.09 & 1.13 & 1.20 & 1.25 & 1.26 & 1.25 & 1.26 \\
Time-MoE Weather & 1.82 & 2.16 & 1.92 & 2.50 & 2.39 & 2.64 & 2.63 \\
TiRex ETTh1 & 2.03 & 1.57 & 1.07 & 0.76 & 0.63 & 0.69 & 0.69 \\
\bottomrule
\end{tabular}
\caption{Batch-size sweeps reporting speculative speedup ($\times$)
against the matched target-only baseline at the same batch size.
Settings are Timer-XL ECL $K{=}3$, $\sigma{=}0.25$; TimesFM ETTh1
$K{=}1$, $\sigma{=}0.10$; Time-MoE Weather $K{=}3$, $\sigma{=}0.15$;
and TiRex ETTh1 $K{=}4$, $\sigma{=}0.20$. MSE at fixed $\sigma$ varies
by less than $0.002$ across batches (TimesFM, Timer-XL). Each sweep is
its own run, not comparable with Table~\ref{tab:main} of the main
paper.}
\label{tab:bsweep}
\end{table}

\begin{table}[tp]
\centering
\small
\setlength{\tabcolsep}{4pt}
\begin{tabular}{@{}l p{4.6cm}@{}}
\toprule
\multicolumn{2}{@{}l}{\emph{Target}} \\
Checkpoint & trained per dataset (this paper) \\
Parameters & 67.4M \\
Architecture & 8 layers, $d_{\text{model}}$ 1024, $d_{\text{ff}}$ 2048, 8 heads \\
Context length & 1536 \\
Patch (in / out per forward) & 96 / 96 \\
Output head & point; instance normalization on \\
Precision & fp32 \\
\midrule
\multicolumn{2}{@{}l}{\emph{Draft}} \\
Source & trained $0.125\times$-width (Table~\ref{tab:trainhparams}) \\
Parameters & 158K ($0.23\%$ of target) \\
Architecture & 1 layer, $d_{\text{model}}$ 128, $d_{\text{ff}}$ 256, 1 head \\
Context length & 1536 \\
\midrule
\multicolumn{2}{@{}l}{\emph{Speculative decoding}} \\
$K$ (reported) & 1--3 \\
$\sigma$ grid (reported) & 0.05--0.35 (0.10--0.35) \\
Proposal & draft mean \\
Fallback on reject & target mean ($r{=}0$) \\
Acceptance draw & Bernoulli, \eqref{eq:logratio} \\
Verification & single pass \\
\midrule
\multicolumn{2}{@{}l}{\emph{Evaluation}} \\
Prediction length & 336 / 720 \\
Batch size (reported) & 1--64 \\
Warm-up batches & 2 \\
\bottomrule
\end{tabular}
\caption{Timer-XL configuration.}
\label{tab:hp-timerxl}
\end{table}

\begin{table}[tp]
\centering
\small
\setlength{\tabcolsep}{4pt}
\begin{tabular}{@{}l p{4.6cm}@{}}
\toprule
\multicolumn{2}{@{}l}{\emph{Target}} \\
Checkpoint & \path{google/timesfm-2.5-200m-pytorch} \\
Parameters & 231.3M \\
Context length & 2048 \\
Patch (in / out per forward) & 32 / 128 \\
Output head & quantiles (median); quantile head on \\
Flip invariance & off in all three modes \\
Precision & fp32 \\
\midrule
\multicolumn{2}{@{}l}{\emph{Draft}} \\
Source & same checkpoint \\
Context length & 256 (swept $\{64,128,256,512\}$) \\
Configuration & quantile head off \\
\midrule
\multicolumn{2}{@{}l}{\emph{Speculative decoding}} \\
$K$ (reported) & 1 \\
$\sigma$ grid (reported) & 0.05--0.35 (0.10--0.30) \\
Proposal & draft mean \\
Fallback on reject & target mean ($r{=}0$) \\
Acceptance draw & Bernoulli, \eqref{eq:logratio} \\
Verification & single pass \\
\midrule
\multicolumn{2}{@{}l}{\emph{Evaluation}} \\
Prediction length & 1536 \\
Batch size (reported) & 4--8 \\
Warm-up batches & 2 \\
\bottomrule
\end{tabular}
\caption{TimesFM 2.5 configuration.}
\label{tab:hp-timesfm}
\end{table}

\begin{table}[tp]
\centering
\small
\setlength{\tabcolsep}{4pt}
\begin{tabular}{@{}l p{4.6cm}@{}}
\toprule
\multicolumn{2}{@{}l}{\emph{Target}} \\
Checkpoint & \path{thuml/sundial-base-128m} \\
Parameters & 128.3M \\
Context length & 1536 \\
Patch (in / out per forward) & 16 / 16 \\
Output head & flow matching, $n_s{=}20$ samples \\
Precision & fp32 (loader default) \\
\midrule
\multicolumn{2}{@{}l}{\emph{Draft}} \\
Source & same checkpoint \\
Context length & 512 (swept $\{256,512,768\}$) \\
Configuration & $n_s{=}1$ \\
\midrule
\multicolumn{2}{@{}l}{\emph{Speculative decoding}} \\
$K$ (reported) & 1 \\
$\sigma$ grid (reported) & 0.05--0.35 (0.15--0.30) \\
Proposal & draft mean \\
Fallback on reject & target mean ($r{=}0$) \\
Acceptance draw & Bernoulli, \eqref{eq:logratio} \\
Verification & single pass, channel-batched; verifier $n_s{=}20$ (matched) \\
\midrule
\multicolumn{2}{@{}l}{\emph{Evaluation}} \\
Prediction length & 336 / 720 \\
Batch size (reported) & 4--128 \\
Warm-up batches & 2 \\
\bottomrule
\end{tabular}
\caption{Sundial configuration.}
\label{tab:hp-sundial}
\end{table}

\begin{table}[tp]
\centering
\small
\setlength{\tabcolsep}{4pt}
\begin{tabular}{@{}l p{4.6cm}@{}}
\toprule
\multicolumn{2}{@{}l}{\emph{Target}} \\
Checkpoint & \path{Maple728/TimeMoE-200M} \\
Parameters & 453.2M \\
Context length & 3072 \\
Patch (in / out per forward) & 1 (point) / 64 (horizon head) \\
Output head & point; multi-resolution horizons $\{1,8,32,64\}$ \\
Precision & bf16 \\
\midrule
\multicolumn{2}{@{}l}{\emph{Draft}} \\
Source & \path{Maple728/TimeMoE-50M} \\
Parameters & 113.4M \\
Context length & 768 (swept $\{512,768,1024\}$) \\
\midrule
\multicolumn{2}{@{}l}{\emph{Speculative decoding}} \\
$K$ (reported) & 3 \\
$\sigma$ grid (reported) & 0.05--0.30 (0.15--0.30) \\
Proposal & draft mean \\
Fallback on reject & target mean ($r{=}0$) \\
Acceptance draw & Bernoulli, \eqref{eq:logratio} \\
Verification & single pass via horizon head \\
\midrule
\multicolumn{2}{@{}l}{\emph{Evaluation}} \\
Prediction length & 1536 \\
Batch size (reported) & 8--128 \\
Warm-up batches & 2 \\
\bottomrule
\end{tabular}
\caption{Time-MoE configuration.}
\label{tab:hp-timemoe}
\end{table}

\begin{table}[tp]
\centering
\small
\setlength{\tabcolsep}{4pt}
\begin{tabular}{@{}l p{4.6cm}@{}}
\toprule
\multicolumn{2}{@{}l}{\emph{Target}} \\
Checkpoint & \path{NX-AI/TiRex} \\
Parameters & 35M \\
Context length & 2048 \\
Patch (in / out per forward) & 32 / $m{\cdot}32$ at rollout $m{=}4$ \\
Output head & 9 quantiles (median) \\
Precision & library default \\
\midrule
\multicolumn{2}{@{}l}{\emph{Draft}} \\
Source & same checkpoint \\
Context length & 2048 \\
Configuration & rollout $m{=}16$ \\
\midrule
\multicolumn{2}{@{}l}{\emph{Speculative decoding}} \\
$K$ (reported) & 4 ($=16/4$) \\
$\sigma$ grid (reported) & 0.05--0.35 (0.05--0.20) \\
Proposal & draft median \\
Fallback on reject & target chunk \\
Acceptance draw & threshold $\norm{\cdot}_d^2\le\sigma^2$, batch-synchronized \\
Verification & batched $K$-prefix call \\
\midrule
\multicolumn{2}{@{}l}{\emph{Evaluation}} \\
Prediction length & 1536 \\
Batch size (reported) & 4 \\
Warm-up batches & 0 \\
\bottomrule
\end{tabular}
\caption{TiRex configuration (standalone implementation).}
\label{tab:hp-tirex}
\end{table}

\begin{table}[tp]
\centering
\small
\setlength{\tabcolsep}{2.5pt}
\begin{tabular}{@{}llp{2.9cm}@{}}
\toprule
 & Distilled draft & Supervised draft \\
\midrule
Datasets & ETTh1 & Weather, ECL, PEMS, Traffic, others \\
Objective & MSE to teacher output & MSE to ground truth \\
Learning rate & $5\times10^{-5}$ & $1\times10^{-4}$ \\
LR schedule & cosine ($T_{\max}{=}10$) & type1 decay \\
Epochs & 15 & 10 \\
Batch size & 64 & 1--64 per dataset \\
\bottomrule
\end{tabular}
\caption{Training hyperparameters of the Timer-XL $0.125\times$ draft
(1 layer, $d_{\text{model}}$ 128, 1 head, $d_{\text{ff}}$ 256; patch 96,
context 1536, instance normalization on). Both variants use weight decay
$0$, dropout $0.1$, seed $2021$, and validation on the last split
(\texttt{valid\_last}). The target's own per-dataset training uses the
supervised recipe at learning rate $1\times10^{-4}$, 10 epochs. Datasets
where the supervised draft matched or beat the target are excluded by the
accuracy gate (main text).}
\label{tab:trainhparams}
\end{table}

\section{Additional Related Work}
\label{app:related}

\paragraph{Speculative Decoding Beyond Discrete Tokens.}
Recent work carries SD into continuous outputs where exactness stays
affordable: \citet{wang2024continuous} accelerate diffusion-head image
autoregression with a continuous acceptance ratio and
acceptance-rejection resampling; for event streams,
TPP-SD \citep{gong2025tppsd} ports the full draft--verify--resample
loop to temporal point processes, and
\citet{bilos2025speculative} derive rejection constants from
piecewise-linear bounds on the conditional intensity. All three operate on
scalar event times or compact visual tokens whose densities are exposed
by construction, through purpose-built or closed-form parametric
decoders. None reaches TSFM patches, where
outputs are high-dimensional, released heads expose no usable density,
and exact correction provably costs more than it saves
(Proposition~\ref{prop:rescost}). For forecasting itself,
Apollo-Forecast \citep{yin2025apollo} quantizes series into discrete
tokens and races draft against target under a tolerance heuristic, with
no output-law analysis and the quantization artifacts our
density-native framework avoids. Our
Chronos-T5 finding (Appendix~\ref{app:implementation}) delineates the two
regimes, with token-level rules for discrete forecasters and continuous
rules for continuous heads.

\ifx\maininclude\undefined
\bibliography{references}

\end{document}
\fi

\end{document}